\documentclass[final]{latex/cvpr}

\usepackage{times}
\usepackage{epsfig}
\usepackage{graphicx}
\usepackage{amsmath}
\usepackage{amssymb}
\usepackage{nopageno}

\usepackage{booktabs}
\usepackage{multirow}
\usepackage{array}
\usepackage{float}

\newcommand{\drule}{\specialrule{0.2pt}{1pt}{1pt}%
            \specialrule{0.2pt}{0pt}{\belowrulesep}%
            }
  {\begin{list}{}%
          {\setlength{\leftmargin}{#1}}%
          \item[]%
  }
  {\end{list}}
  
\newcommand*{\affaddr}[1]{#1} % No op here. Customize it for different styles.
\newcommand*{\affmark}[1][*]{\textsuperscript{#1}}

\usepackage{kotex}

\newcommand{\todow}[1]{\textcolor{red}{#1}}

\newcommand\blfootnote[1]{%
  \begingroup
  \renewcommand\thefootnote{}\footnote{#1}%
  \addtocounter{footnote}{-1}%
  \endgroup
}

\usepackage[pagebackref=true,breaklinks=true,colorlinks,bookmarks=false]{hyperref}

\def\cvprPaperID{0872} % *** Enter the CVPR Paper ID here
\def\confYear{CVPR 2021}
\begin{document}

%%%%%%%%% TITLE
\title{RobustNet: Improving Domain Generalization in Urban-Scene Segmentation\\via Instance Selective Whitening}
% \title{Improving Robustness to Unseen Domains by Instance Selective Whitening Loss\\with Disentangling Covariance}

\author{
{Sungha Choi}$^\text{*}$\affmark[1,3]\,
{Sanghun Jung}$^\text{*}$\affmark[2]\,
{Huiwon Yun}\affmark[4]\,
{Joanne T. Kim}\affmark[3]
\\
{Seungryong Kim}\affmark[3]\,
{Jaegul Choo}\affmark[2]
\vspace*{0.3cm}
\\
\affaddr\affmark[1]LG AI Research\,
\affmark[2]KAIST\,
\affmark[3]Korea University\,
\affmark[4]Sogang University\\
\vspace*{-0.5cm}
% \footnotesize{\email{sungha.choi@lgresearch.ai}\quad\email{shjung13@kaist.ac.kr}\quad\email{vanche9@gmail.com}}
% \vspace*{-0.07cm}
% \\
% \footnotesize{\email{tengyee@korea.ac.kr}\quad\email{seungryong\_kim@korea.ac.kr}\quad\email{jchoo@kaist.ac.kr}}\\
}

\makeatletter
\g@addto@macro\@maketitle{
  \begin{figure}[H]
  \vspace{-0.5cm}
  \setlength{\linewidth}{\textwidth}
  \setlength{\hsize}{\textwidth}
  \centering
  \includegraphics[width=\linewidth]{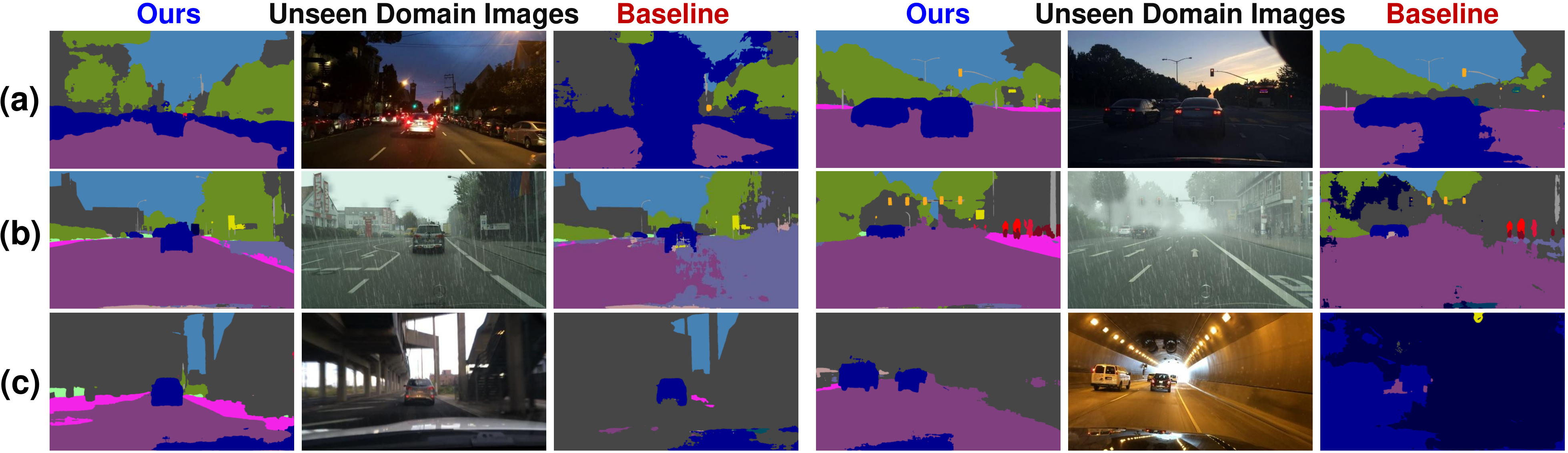}
%   \rule{10cm}{5cm} % this is your image
  \vspace{-0.3cm}
  \caption{Segmentation results on \emph{unseen} domains (\textit{i.e.,} BDD-100K~\cite{yu2020bdd100k} and RainCityscapes~\cite{hu2019depth}) with the models trained on Cityscapes~\cite{Cordts2016Cityscapes}. Note that Cityscapes does not contain the following types of images:
  %These types of images are not seen in the training domain.
  (a) low-illuminated, (b) rainy, and (c) unexpected scenes. Our method makes reasonable predictions in these three cases, while the baseline~\cite{chen2018encoder} model completely fails on them.} 
  %\tr{b랑 c같은 데이터도 training에 없나요?! 그렇다면 이 설명을 앞으로 빼도 좋을 듯해서요!}
%   (b) The segmentation results for rainy images. (c) The segmentation results for unexpected scenes.}
%   Theses conditions are not found in 
  \label{fig:supp_main}
  \vspace{-0.0cm}
  \end{figure}
}
\makeatother

\maketitle
\blfootnote{\hspace*{-0.15cm}\vspace{0.0cm}* indicates equal contribution}

%%%%%%%%% ABSTRACT
\vspace{-0.4cm}
\begin{abstract}
\vspace{-0.2cm}
   Enhancing the generalization capability of deep neural networks to unseen domains is crucial for safety-critical applications in the real world such as autonomous driving.
   To address this issue, this paper proposes a novel instance selective whitening loss to improve the robustness of the segmentation networks for unseen domains. 
   Our approach disentangles the domain-specific style and domain-invariant content encoded in higher-order statistics (\textit{i.e.,} feature covariance) of the feature representations and selectively removes only the style information causing domain shift. As shown in Fig.~\ref{fig:supp_main}, our method provides reasonable predictions for (a) low-illuminated, (b) rainy, and (c) unseen structures. These types of images are not included in the training dataset, where the baseline shows a significant performance drop, contrary to ours.
   Being simple yet effective, our approach improves the robustness of various backbone networks without additional computational cost. 
   We conduct extensive experiments in urban-scene segmentation and show the superiority of our approach to existing work. Our code is available at this link\footnote{\vspace{-0.25cm} \url{https://github.com/shachoi/RobustNet}.}.
%   We will publicly release the code and pre-trained models.
\vspace{-0.4cm}
\end{abstract}
   %The proposed loss constrains the covariance matrix of the feature representation to be close to the identity matrix so as to learn domain-invariant features rather than domain-specific ones.
   %by removing the domain-specific style information in higher-order statistics and enhancing content encoding capability with only negligible architectural changes.
   %\choi{selectively covariance 를 없애는 것을 주요 내용로 업데이트 예정}
   %for autonomous driving in the field of safety-critical applications. and show the superiority of our approach to other existing work.
   %\sh{위의 various backbone network에서도 잘 동작함을 여기서 같이 애기해주는것도 좋을 것 같아요, 실험 부분에 들어가도 괜찮다면요!}   
%%%%%%%%% BODY TEXT
% \choi{현재 (11/15 24:00) 완성된 부분: Introduction, Related Work, Preliminaries, Proposed Method ~ ISW 까지}

\vspace{-0.1cm}
\section{Introduction}\label{introduction}
\vspace{-0.05cm}
When deploying deep neural networks (DNNs) trained on a \emph{given} dataset (\textit{i.e.,} source domain) in real-world \emph{unseen} data (\textit{i.e.,} target domain), DNNs often fail to perform properly due to the domain shift. Overcoming this issue is crucial, especially for safety-critical applications such as autonomous driving. In particular, real-world data consist of unexpected and unseen samples, for example, those images taken under diverse illumination, adverse weather conditions, or from different locations. It is generally impossible to model such a full data distribution with limited training data, so reducing the domain gap between source and target domains has been a long-standing problem in computer vision.

%When deploying deep neural networks (DNNs) trained from a given datasets (\textit{i.e.,} source domain) to the real world (\textit{i.e.,} target domain), can we trust that the DNNs will work reliably? \choi{문장 업데이트 필요}\tr{생각해보고 이 문단 다시 돌아올게요!}
%Finding a solution to this question is crucial, especially for safety-critical applications such as autonomous driving.
%Even if the real world includes significantly different unseen data, which can be caused by diverse illumination, adverse weather conditions, and locational differences, 
%we can only collect a small portion of them.
%Even if the training dataset is large-scale, it is just a small portion of the real world.
%Also, the real world includes significantly different unseen data, which may be caused by diverse illumination, adverse weather conditions, locational differences, and so on.
% Moreover, the real world has an uncertainty that could lead to an unexpected new scene.
%Moreover, in the real world, an unexpected new scene can occur.
%Hence, fully representing the real world with the training dataset is practically impossible. To alleviate these issues, it is essential to reduce the large domain gap between the training data and the real world.

\begin{figure}[t!]
\begin{center}
  \includegraphics[width=1.00\linewidth]{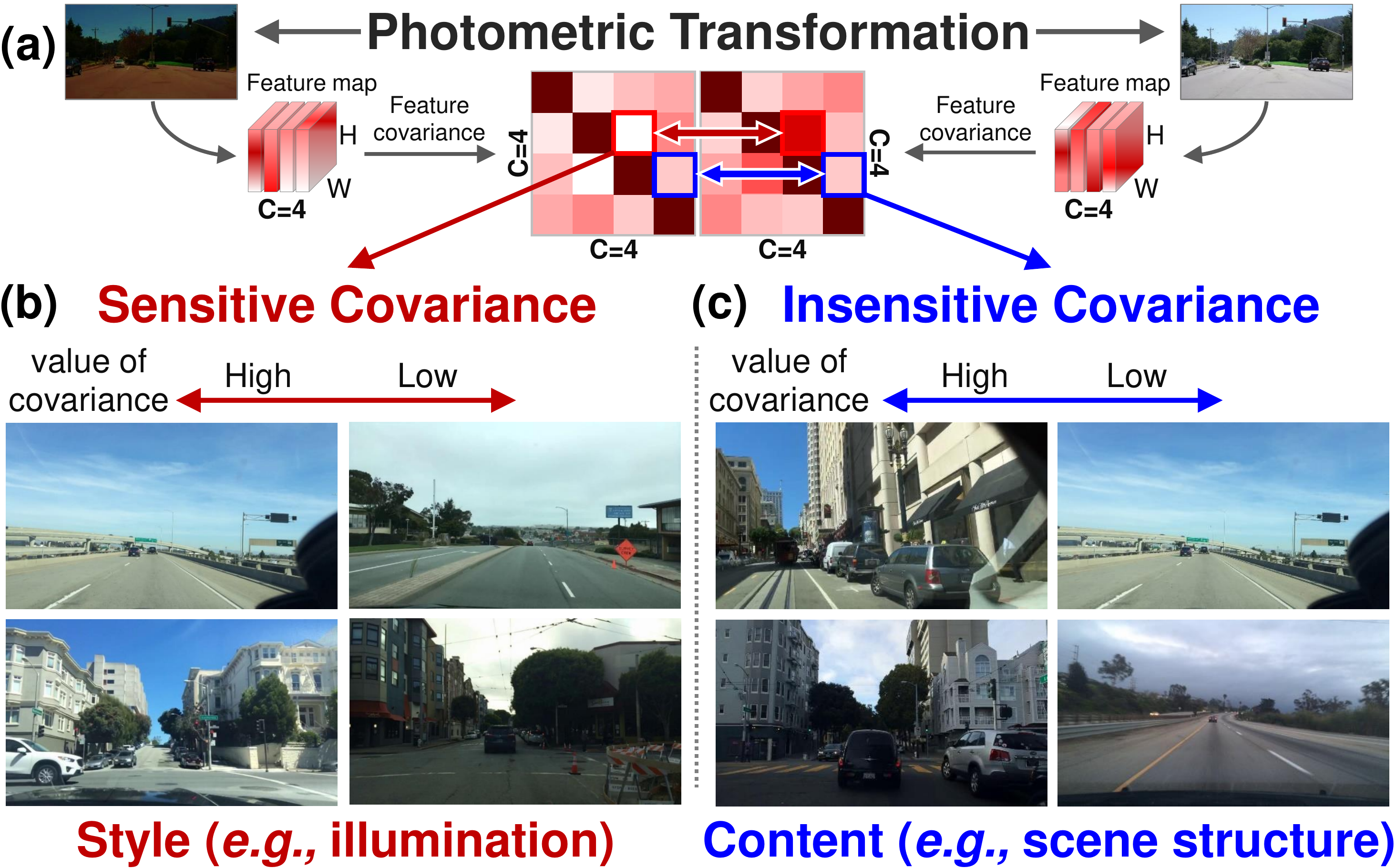}
\end{center}
\vspace*{-0.1cm}
   \caption{\textbf{Overview of our motivation.} (a) We first identify the feature covariance sensitive to the photometric transformation and examine the tendency of the images in each group.
   (b) Sensitive covariances: Illumination (\textit{i.e.,} style) tends to significantly vary.
   (c) Insensitive covariances: Sensitive to scene structure differences (\textit{i.e.,} content) but unaffected by the photometric transformation.
%   We investigate the trend of the images according to the value of the selected covariances (\textit{i.e.,} sensitive and insensitive) from the feature covariance of each image. Looking at the tendency of the images with each covariance value, (b) the illumination (\textit{i.e.,} style) and (c) scene complexity (\textit{i.e.,} content) tend to vary. 
%   The covariance in (c) does not change for the color transformation, but it is sensitive according to the scene complexity. 
Accordingly, we aim to \emph{selectively} remove only the style-sensitive covariances that may cause the domain shift.}
%   the order of the channels are changed by clustering.
%   Y 축은 height 이고 X축은 채널입니다. 즉 각 vertical position 마다 어떤 channel 을 attention 하고 있는지를 나타냄. Note that 여기서의 채널 순서는 보다 나은 보기를 위해서, 비슷한 채널 들끼리 clustering 하였습니다.}
\label{fig:overview}
\vspace*{-0.45cm}
\end{figure}

% One approach to mitigate these problems is domain adaptation (DA)~\cite{ben2007analysis,ganin2016domain,hoffman2018cycada,ganin2015unsupervised,zou2018unsupervised,murez2018image,vu2019advent,pan2020unsupervised,saito2018maximum}. 
Domain adaptation (DA) is an approach to mitigate the performance degradation caused by such a domain gap~\cite{ben2007analysis,ganin2016domain,hoffman2018cycada,ganin2015unsupervised,zou2018unsupervised,murez2018image,vu2019advent,pan2020unsupervised,saito2018maximum}.
Generally, DA focuses on adapting the source domain distribution to that of the target domain, but it requires access to the samples in the target domain, which limits their applicability. 
%When applying the DNNs trained on the partial measurements to the real-world, we cannot even expect the data distribution, as well as access the samples.
When we set the entire real world as a target domain, it is difficult in pactice to obtain data samples that fully cover the target domain.
%the input distributions, which is known as domain shift, requiring access to the data samples in the target domain during training. Even unsupervised domain adaptation relies on unlabeled data samples from the target domain.
%However, when we set the real world as the target domain, it is difficult to obtain images that could fully represent the target domain due to its unpredictability.
% it is difficult even to obtain images that represent the target domain fully since the real world is infinitely large and unpredictable. 

% Therefore, one needs a method to provide the robustness to DNNs on arbitrary unseen domains, which cannot be addressed by the domain adaptation methods. To tackle down these issues, domain generalization (DG)~\cite{muandet2013domain} has been proposed to increase the robustness to unseen domains without access to any data in the target domain during training. 

Domain generalization (DG) overcomes this limitation by improving the robustness of DNNs to arbitrary unseen domains. 
In general, most DG methods~\cite{li2018domain,seo2019learning,dou2019domain,motiian2017unified,muandet2013domain,Ghifary_2015_ICCV,li2017learning,balaji2018metareg,li2019episodic,li2019feature} accomplish this through the learning of a shared representation across multiple source domains. However, collecting such multi-domain datasets is costly and labor-intensive, and furthermore, the performance highly depends on the number of source datasets.
A recent study~\cite{pan2018two} has shown that the DG problem can be addressed by exploiting instance normalization layers~\cite{ulyanov2016instance} instead of relying on multiple source domains, leading to a simple and cost-effective training process. The instance normalization just standardizes features while not considering the correlation between channels.
However, a number of studies~\cite{gatys2015texture,gatys2016image,li2017universal,siarohin2018whitening,roy2019unsupervised,luo2017learning,cho2019image,pan2019switchable,sun2016deep} claim that feature covariance contains domain-specific style such as texture and color. This implies that applying instance normalization to the networks may not be sufficient for domain generalization, because the feature covariance is not considered.
A whitening transformation is a technique that removes feature correlation and makes each feature have unit variance. It has been proven that the feature whitening effectively eliminates domain-specific style information as shown in image translation~\cite{cho2019image}, style transfer~\cite{li2017universal}, and domain adaptation~\cite{pan2019switchable,sun2016deep,roy2019unsupervised}, and thus it may improve the generalization ability of the feature representation, but not yet fully explored in DG.
% but (add limitation here!). Another possible approach to normalizing the feature is a whitening normalization~\cite{?} that \sr{(add description for WN.)}, which has been proven that it effectively eliminates domain-specific style in style transfer literature~\cite{?} and thus may improve the generalization ability of the feature representation, but not fully explored in domain generalization yet.
%However, another normalization method, whitening, is less explored in DG. 
%The whitening is a technique that can manage not only standardizing but also decorrelating features.
% However they do not consider the pair-wise feature covariance. 
%A number of studies~\cite{gatys2015texture,gatys2016image,li2017universal,siarohin2018whitening,roy2019unsupervised,luo2017learning,cho2019image,pan2019switchable,sun2016deep} present that feature covariance contains domain-specific style such as texture and color. 
% \tr{an image 의미 모르겟서요..}\choi{Got it, image-specific 이였는데 사실 필요없어요}
%Therefore, whitening the feature map, which eliminates the feature correlation, can remove domain-specific style and has the potential to improve the generalization performance.
However, simply adopting the whitening transformation to improve the robustness of DNNs is not straightforward, since it may eliminate domain-specific style and domain-invariant content at the same time. Decoupling the two factors and selectively removing the domain-specific style is the main scope of this paper.
%First, if we remove the correlation between channels completely, this may result in poor performance, because the domain-invariant content (such as shape) can be lost along with the domain-specific style (such as color and texture).
%Specifically, whitening transformation diminishes the discriminative power of features~\cite{pan2019switchable,wadia2020whitening} and distorts the boundary of the objects~\cite{li2018closed,li2017universal}. 
% \tr{여기까지는 selectiveness이야기하고,} \tr{여기에서는 기존 whitening에 드는 cost는 큰데 우리 method는 가벼울 것을 이야기하고 싶은것 맞나요} \choi{맞습니다. 의견을 살려서, selectiveness + cost 에 대한 문제제기 그리고 아랫부분에 문제해결을 위해 selectiveness 에다가 cost 개선 추가하였습니다}
%Another concern is that whitening transformation is computationally expensive~\cite{huang2018decorrelated,pan2019switchable,cho2019image}. %, and removes domain-invariant content features as well as domain-specific style~\cite{li2017universal}.
% In order to make good use of the whitening for solving the DG problem, we 
%Consequently, a new manner to make good use of the whitening, which overcomes these limitations, is required to solve the DG problem.
% whitening method that overcomes these limitations is required to solve DG problem.

In this paper, we present an instance \emph{selective} whitening loss that alleviates the limitations of the existing whitening transformation for domain generalization, by selectively removing information that causes a domain shift while maintaining a discriminative power of feature within DNNs. Our method does not rely on an explicit \emph{closed-form} whitening transformation, but implicitly encourage the networks to learn such a whitening transformation through the proposed loss function, thus requiring negligible computational cost.
As illustrated in Fig.~\ref{fig:overview}, our method selectively removes only those feature covariances that respond sensitively to photometric augmentation such as color transformation.
Our experiments on urban-scene segmentation in DG settings, performed using several backbone networks, show evidence that our approach consistently boosts the DG performance.
The main contributions include the following:
\vspace{-0.0cm}
\begin{itemize}
\vspace{-0.2cm}
    \item We propose an instance selective whitening loss for domain generalization, which disentangles domain-specific and domain-invariant properties from higher-order statistics of the feature representation and  selectively suppresses domain-specific ones.
\vspace{-0.2cm}
    \item Our proposed loss can easily be used in existing models and significantly improves the generalization ability with negligible computational cost.
\vspace{-0.2cm}
    \item We apply the proposed loss to urban-scene segmentation in a DG setting and show the superiority of our approach over existing approaches in both a qualitative and quantitative manner.
\end{itemize}

\section{Related Work}
%When out-of-distribution data is given to the DNNs, it suffers from the degradation of performance. This problem is caused by the domain shift or distribution discrepancy.
% \paragraph{Domain adaptation and generalization}
% \tr{제가 이해한 바로는 저희 method는 domain generalization에 속하는거같은데 인트로에서 말한것 외에 따로 언급해야할게 없다면 da는 related work에서 빼는게 어떨까욥}\out{DA셋팅으로 따로 실험 비교가 없다면 빼는것도 괜찮을것같네요}
% TO DO
% DA methods tackle the problem of domain gap between the source and the target domains
% while data is labeled in the source domain only.
% There have been considerable studies about DA.
% Common approach is to minimize the feature distance loss directly~\cite{sun2016deep,long2015learning,peng2019moment} or distinguish the domain-invariant features from domain-specific features []. Some works used adversarial learning~\cite{tzeng2017adversarial,sankaranarayanan2018generate,saito2018maximum}.
\vspace{-0.15cm}
\paragraph{Domain adaptation and generalization}
% \out{DA/DG 에 대한 연구 소개. DA 보다 DG 를 더 Focusing 함. DG 방법론에 대한 분류 (e.g. meta learning, normalization, randomization, 등등\\}
It is well known that significant labeling efforts are required so as to ensure the reliable performance of various tasks such as semantic segmentation~\cite{long2015fully,badrinarayanan2017segnet,chen2017rethinking,zhu2019improving,choi2020cars}. To tackle this challenge, domain adaptation (DA) methods were proposed to transfer the knowledge learned from abundant labeled data (\textit{i.e.,} a source domain) to a target domain where labeled data are scarce. In contrast to DA, domain generalization (DG) methods assume that the model cannot access the target domain during training and aim to improve the generalization ability to perform well in an unseen target domain. Various approaches such as meta-learning~\cite{li2017learning,balaji2018metareg,li2019episodic,li2019feature}, 
adversarial training~\cite{li2018domain,li2018deep,rahman2020correlation}, autoencoder~\cite{Ghifary_2015_ICCV,li2018domain}, metric learning~\cite{dou2019domain,motiian2017unified}, 
data augmentation~\cite{yue2019domain,gong2019dlow,zhou2020learning} have been proposed to learn domain-agnostic feature representations.
% One of main approaches is based on data augmentation. CNN-based networks can easily be biased on texture~\cite{geirhos2019imagenettrained} which means models trained by source domain data can be biased towards source domain’s texture. Therefore, images are augmented into various styles to prevent overfitting on domain-specific styles and learn domain-invariant visual representation~\cite{zhou2020learning,yue2019domain,gong2019dlow}. Another Approach is to align features of both source and target domain to same embedding space via adversarial learning~\cite{li2018domain,rahman2020correlation,li2018deep}. Several methods~\cite{balaji2018metareg,li2017learning,dou2019domain} adopted a meta learning scheme to DG. Since these methods are task-agnostic, they~\cite{balaji2018metareg,li2017learning} showed their method performed well on both vision and nlp tasks. ~\cite{chattopadhyay2020learning} utilized both domain-specific and domain-invariant features in a multi source domain setting. They insisted features which have domain characteristics can improve model performance. 
Recently, several studies~\cite{pan2018two,seo2019learning} have shown the effectiveness of exploiting both batch normalization (BN)~\cite{ioffe2015batch} and instance normalization (IN)~\cite{ulyanov2016instance} within DNNs to solve the DG problem. These studies show that BN improves discriminative ability on features, while IN prevents overfitting on training data, so that generalization performance is improved on unseen domains by combining BN and IN. Especially, IBN-Net~\cite{pan2018two} shows a significant performance improvement with the marginal architectural modification that incorporates the IN layers through training on a single source domain, unlike most DG methods that require multiple source domains. 
%Our work is related to IBN-Net exploiting normalization layer.
This normalization based DG method is attractive because it can be applied as a complement to other DG methods based on multiple source domains.% relying on multiple source domain.
% these methods~\cite{pan2018two,seo2019learning}.
% IBN-Net~\cite{pan2018two}, DSON~\cite{seo2019learning}, SW~\cite{pan2019switchable} used normalization techniques such as BN and IN. BN~\cite{ioffe2015batch} normalizes features within mini-batch which helps optimization. IN~\cite{ulyanov2016instance} can remove style of image so that it is usually used in style transfer. IBN-net~\cite{pan2018two} designed architecture using them and DSON~\cite{seo2019learning} combined them. Our work is related to these methods~\cite{pan2018two,seo2019learning}. However, we use IW instead of IN which removes domain-specific information effectively. Furthermore we devised instance selective whitening to remove more domain-specific information.
% \out{Normalization 과 우리 방법론과의 연결고리 언급}
\vspace{-0.45cm}
\paragraph{Semantic segmentation in DG}
%\out{Semantic segmentation 분야에서의 DG focusing}
%\out{연구가 덜 되었다는 내용 포함}
%\choi{DA for segmentaton 은 연구가 많이 되고 있다고 언급하고 넘어가죠. 특히 GTA, Synthia 등의 Synthetic data 로 인해서, DA 가 많이 연구되고 있다고}
Based on the synthetic data such as GTAV~\cite{Richter_2016_ECCV} and SYNTHIA~\cite{Ros_2016_CVPR}, numerous DA studies~\cite{pan2020unsupervised,vu2019advent,saleh2018effective,chen2018road,zou2018unsupervised,hoffman2018cycada,tsai2018learning,ma2018exemplar,zhang2017curriculum} have been proposed in semantic segmentation, but only a few DG studies~\cite{yue2019domain,pan2018two} address semantic segmentation, as the majority of the DG methods mainly focused on image classification.
% Numerous previous works for DA and DG have focused on image classification, using large scale synthetic dataset such as GTAV~\cite{Richter_2016_ECCV}, SYNTHIA~\cite{Ros_2016_CVPR} in training time. 
% DA methods for semantic segmentation have begun to appear~\cite{murez2018image,hoffman2018cycada,tsai2019domain,tsai2018learning,vu2019advent,zou2018unsupervised,zou2020confidence,yue2019domain,pan2018two}. Especially large scale synthetic dataset such as GTAV~\cite{Richter_2016_ECCV}, SYNTHIA~\cite{Ros_2016_CVPR} was used in training time. % Though DA methods show better performance in general, DG methods is considered more useful in the real-world application for that target domain is not required during training.
% While DA is \tr{성능값이 DA가 더 좋다는 의미인가요!?}\yun{넵, 수정이 필요할까요??}\tr{제가 이해가 안되어서 여쭤봤어요ㅎㅎ 감사합니다!}superior since it has the advantage of accessing target domain, DG is more useful in real application. 
DA, which can access the target domains, generally has better performance than DG, but DG methods that can handle an arbitrary unseen domain without access to the target domain are mandatory in the real world. This paper focuses on the DG method practically helpful in semantic segmentation where various conditions exist such as adverse weather, diverse illumination, location differences, and so on.

% In practice, there are limitations collecting data about all possible situations while out-of-distribution can harm model performance~\cite{recht2018cifar10,sankaranarayanan2018learning,saleh2018effective}.
% Previous DG approaches mainly utilized data augmentation in style~\cite{gong2019dlow,yue2019domain}, while other method~\cite{zhang2020generalizable} exploited normalization. In this work, we also suggest to exploit normalization for semantic segmentation. 
% Our method is also included in this approach. %\out{ibn net이 dg 실험을 안해서 여기에 넣으면 안됨... 문장자체 수정필요할수도}
% \paragraph{Semantic Segmentation in Domain Adaptation}
% \paragraph{Semantic Segmentation in Domain Generalization}
\vspace{-0.45cm}
\paragraph{Feature covariance}
%\out{채널 covariance 에 대한 사전 연구 들}
The seminal studies~\cite{gatys2015texture,gatys2016image} have demonstrated that feature correlations (\textit{i.e.,} a gram matrix or covariance matrix) take style information of images. Since then, numerous studies exploit the feature correlation in style transfer~\cite{li2017universal}, image-to-image translation~\cite{cho2019image}, domain adaptation~\cite{roy2019unsupervised,sun2016deep} and networks architecture~\cite{luo2017learning,pan2019switchable,huang2018decorrelated,siarohin2018whitening}. Especially, the whitening transformation that removes feature correlation and makes each feature have unit variance, has been known to help to remove the style information from the feature representations~\cite{li2017universal,pan2019switchable,cho2019image}. Our work explores the whitening transformation to improve domain generalization performance. To the best of our knowledge, this is the first attempt to apply whitening to DG.
% Note that GDWCT~\cite{cho2019image} uses the zero-centered feature to compute the covariance matrix, but our method differs in using the standardized feature.
% By using covariance matrix, WCT~\cite{li2017universal} proves that style can be transferred while retaining content information in effortless manner. GDWCT~\cite{cho2019image} extended WCT and successfully reduced the complexity. 
% On the other hand, DBN~\cite{huang2018decorrelated} and SW~\cite{pan2019switchable} focused on that whitening can be used as a general technique for optimization. DBN~\cite{huang2018decorrelated} showed increased generalization ability and fast convergence by applying whitening within mini-batch. SW~\cite{pan2019switchable} combined several normalizations including whitening and standardizations without manual adjustment. 
% In this paper, we exploit whitening in DG to discard domain-specific style information \tr{successively}. To the best of our knowledge, this is the first attempt to use whitening in DG. 
% Besides, previous studies adopting whitening transformation for DA have drawbacks with respect to complexity. We deal with this issue by using whitening transformation based on loss.
% Although there are methods which used whitening in DA~\cite{sun2016return,roy2020unsupervised}, these methods differ from our method in that they use whitening transformation which have drawbacks of expensive expensive complexity. On the contrary we use whitening transformation based on loss.

\begin{figure*}[ht!]
\vspace*{-0.0cm}
  \centering\includegraphics[width=0.97\linewidth]{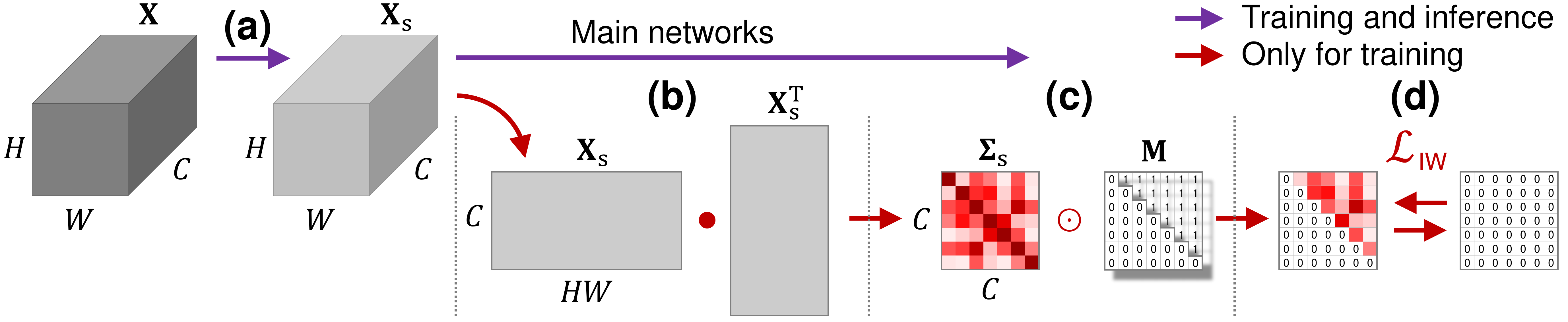}
  \vspace*{-0.0cm}
  \caption{\textbf{Overall process of our proposed method.} (a) Instance standardization. (b) Deriving a covariance matrix from a standardized feature map. (c) Leaving only the covariance to which the whitening loss is applied. 
  (d) Applying the criterion that measures the mean absolute error between the remaining covariance values and zero. No additional computation is required for inference as the operations in red are used only for training.
  Notations; $\mathbf{X}$: intermediate feature map, $\mathbf{X_s}$: standardized feature map, $\mathbf{\Sigma_s}$: covariance matrix of the standardized feature map, $\mathbf{M}$: matrix for masking, $\mathcal{L}_{\text{IW}}$: our proposed instance whitening loss.
  %   Matrix (bold) notations stand for feature maps;
  %   \todo{Feature map notations;}
%   Each operation $op$ is notated as $\text{G}_{op}$, and feature maps are in bold--$\text{X}_{\ell}$: lower-level feature map, Z: width-wise pooled $\text{X}_{\ell}$, $\hat{\text{Z}}$: down-sampled Z, $\text{Q}^{n}$: $n$-th intermediate feature map of 1D convolution layers, $\hat{\text{A}}$: down-sampled attention map, A: final attention map, $\text{X}_{h}$: higher-level feature map, $\Tilde{\text{X}}_{h}$: transformed new feature map. Details can be found in Section~\ref{sec:method_block}.
}
\label{fig:whitening_overview}
\vspace*{-0.4cm}
\end{figure*}

\vspace*{-0.05cm}
\section{Preliminaries}
\vspace*{-0.1cm}
\paragraph{Whitening transformation (WT)} Let $\mathbf{X}\in\mathbb{R}^{C\times HW}$ denote the intermediate feature map, where $C$ is the number of channels, $H$ and $W$ are the spatial dimensions of the feature map, height and width, respectively.\vspace{0.02cm}
WT is a linear transformation that makes the variance term of each channel equal to one and the covariances between each pair of channels equal to zero.
% rearranges the input data to unit-sphere. 
A whitening-transformed feature map\vspace{0.05cm}
$\mathbf{\tilde{X}}$ from $\mathbf{X}$ satisfies that $\mathbf{\tilde{X}}\cdot\mathbf{\tilde{X}}^\top=(HW)\cdot\mathbf{I}\in\mathbb{R}^{C\times C}$, where $\mathbf{I}$ denotes the identity matrix, and can be computed as
\begin{equation} \label{eq_whitening_transform}
\mathbf{\tilde{X}}=\mathbf{\Sigma}_{\mu}^{-\frac{1}{2}}\left(\mathbf{X}-\boldsymbol{\mu}\cdot{\mathbf{1}^\top}\right),
\end{equation}
where $\mathbf{1}\in\mathbb{R}^{HW}$ is a column vector of ones, and $\boldsymbol{\mu}$ and $\mathbf{\Sigma}_{\mu}$ are the mean vector and the covariance matrix, respectively, \textit{i.e.,} 
\vspace{-0.1cm}
\begin{equation}\label{eq_mean}
\vspace{-0.05cm}
\boldsymbol{\mu}=\tfrac{1}{HW}\mathbf{X}\cdot{\mathbf{1}}\in \mathbb{R}^{C\times{1}},
\end{equation}
\begin{equation}\label{eq_covariance}
\mathbf{\Sigma}_{\mu}=\tfrac{1}{HW}\left(\mathbf{X}-\boldsymbol{\mu}\cdot{\mathbf{1}^\top}\right)\left(\mathbf{X}-\boldsymbol{\mu}\cdot{\mathbf{1}^\top}\right)^\top\in\mathbb{R}^{C\times{C}}.
\end{equation}
Since the covariance matrix $\mathbf{\Sigma}_{\mu}$ can be further eigen-decomposed such that $\mathbf{Q}\mathbf{\Lambda}\mathbf{Q}^\top$, where $\mathbf{Q}\in\mathbb{R}^{C\times{C}}$ is the orthogonal matrix of eigenvectors, and $\mathbf{\Lambda}\in\mathbb{R}^{C\times{C}}$ is the diagonal matrix that contains each eigenvalue of the corresponding eigenvector from $\mathbf{Q}$, %From decomposed $\mathbf{\Sigma}_{\mu}$
we can calculate an inverse square root of the covariance matrix $\mathbf{\Sigma}_{\mu}^{-\frac{1}{2}}$ as
\vspace{-0.05cm}
\begin{equation}\label{eq_inverse_square_root}
\mathbf{\Sigma}_{\mu}^{-\frac{1}{2}}=\mathbf{Q}\mathbf{\Lambda}^{-\frac{1}{2}}\mathbf{Q}^\top.
\vspace{-0.05cm}
\end{equation}
% It has been known that WT can increase optimization efficiency~\cite{huang2018decorrelated} when applied to mini-batch in image classification, and can remove style information by applying to each instance in style transfer~\cite{li2017universal}.
It has been known that WT can effectively remove style information by being applied to each instance in style transfer~\cite{li2017universal}.

\vspace{-0.4cm}
\paragraph{Limitations of WT}
We can compute the whitening transformation matrix $\mathbf{\Sigma}_\mu^{-\frac{1}{2}}$ analytically through Eq.~\eqref{eq_inverse_square_root}, but eigenvalue decomposition is computationally expensive, leading to slow training and inference speed 
% and the gradients are difficult to be back-propagated~\cite{huang2018decorrelated, cho2019image}.
and prevents the gradient back-propagation~\cite{huang2018decorrelated, cho2019image}.
To alleviate these problems, previous studies have shown that the goal of WT can be achieved without the eigen-decomposition through the whitening loss~\cite{cho2019image} or approximating the whitening transformation matrix using Newton's iteration~\cite{huang2019iterative,huang2018decorrelated,pan2019switchable}.

Especially, GDWCT~\cite{cho2019image} proposes the deep whitening transformation (DWT) that implicitly makes the covariance matrix $\mathbf{\Sigma}_{\mu}$ close to the identity matrix $\mathbf{I}$ by means of %, which can replace the WT as a closed-form solution.
the loss %for the deep whitening transformation is
defined as
\vspace{-0.1cm}
\begin{equation} \label{eq_whitening_loss}
\mathcal{L}_{\text{DWT}} = \mathbb{E}[\Vert\mathbf{\Sigma_\mu} - \mathbf{I}\Vert_1],
\vspace{-0.1cm}
\end{equation}
where $\mathbb{E}$ denotes the arithmetic mean.
GDWCT applies this loss to image-to-image translation for more significant style changes than other methods~\cite{huang2018multimodal,lee2018diverse} of aligning only the first-order statistics (\textit{i.e.,} channel-wise mean and variance).
% matching only channel-wise statistics, the mean and variance, not the covariance matrix.
% On the other hand, several approaches~\cite{huang2018decorrelated, pan2019switchable} attempt to overcome these drawbacks (\textit{i.e.,} expensive time complexity and non-trivial backpropagation) by approximating the whitening transformation matrix using newton's iteration suggested in IterNorm~\cite{huang2019iterative}.
% Previous study~\cite{cho2019image} have shown that this goal can be achieved through the whitening loss (Eq.~\eqref{eq_whitening_loss}) without the closed form solution (Eq.~\eqref{eq_inverse_square_root}). 
However, applying these alternative methods of WT to DG is not straightforward. Whitening all covariance elements may diminish feature discrimination~\cite{pan2019switchable,wadia2020whitening} and distort the boundary of an object~\cite{li2018closed,li2017universal} because domain-specific style and domain-invariant content are simultaneously encoded in the covariance of the feature map.

\vspace{-0.0cm}
\section{Proposed Method}\label{sec:proposed_method}
\vspace{-0.05cm}
% \sh{Introduction, Related work 에서의 용어들과 통일 필요. 현재는 Instance Whitening, whitening loss, deep instance whitening, deep whitening loss 네개가 쓰이고있는듯 합니다}
% % brief summary of our approaches; two folds - Selective Instance Whitening and Colorizing by multi task learning
% \out{Method section 에 대한 흐름 소개. model overview - whitening loss (orthogonal) - (Budget) - (Selective) - (Something)}
This section presents our approach to solve the domain generalization problem through whitening the feature representation by mitigating undesirable effects of a whitening transformation.
Our method disentangles the covariance into the encoded style and content so that only the style information can be selectively removed, thus increasing the domain generalization ability. We firstly propose an instance whitening and instance-relaxed loss in Section~\ref{method:deep_instance_whitening} and then finally propose our novel instance selective whitening loss in Section~\ref{method:deep_instance_selective_whitening}.
\vspace{-0.05cm}
\subsection{Instance Whitening Loss}\label{method:deep_instance_whitening}
\vspace{-0.1cm}
This subsection describes a series of steps to transform the input feature into the whitening transformed feature as shown in Fig.~\ref{fig:whitening_overview}. Note that our method is applied to each instance, not to a mini-batch.
% \vspace{-0.2cm}
% \paragraph{Standardization (Fig.~\ref{fig:whitening_overview}(a, b))}
% \tr{$\mathbf{\Sigma}_{\mu\,(i,i)}$나 $\mathbf{\Sigma}_{\mu}^{i,i}$, $\mathbf{\Sigma}_{\mu}^{(i,i)}$ 이런식으로 바꾸는건 어떨까요?! i,j가 좀 더 구분이 잘 될거같아요}\choi{Good Idea!!}
Let $\mathbf{\Sigma}_{\mu\,(i,i)}$ denote a diagonal element ($i,i$) and $\mathbf{\Sigma}_{\mu\,(i,j)}$ denote an off-diagonal element ($i,j$) of the covariance matrix $\mathbf{\Sigma}_{\mu}$ of the intermediate feature map, where $0\le i,j<C$, $i\ne j$.
The DWT loss in Eq.~\eqref{eq_whitening_loss} 
%, proposed in GDWCT~\cite{cho2019image} to transform the covariance matrix to identity matrix,
can be decomposed as
%{기존 방법임을 명시하자, 5번은 diagonal 에 6번은 off-diagonal 에 걸리는 loss 임을 명시하자}
\vspace{-0.05cm}
\begin{equation}\label{eq_cov_elements_loss1}
{\big\Vert}\mathbf{\Sigma}_{\mu\,(i,i)}-1{\big\Vert}_1={\big\Vert}\tfrac{\mathbf{x}_{i}^\top\cdot\mathbf{x}_{i}}{HW}-1{\big\Vert}_1={\big\Vert}\tfrac{\vert\mathbf{x}_{i}\vert\vert\mathbf{x}_{i}\vert\cos\text{0}^{\circ}}{HW}-1{\big\Vert}_1
\vspace{-0.05cm}
\end{equation}
\begin{equation}\label{eq_cov_elements_loss2}
\Vert\mathbf{\Sigma}_{\mu\,(i,j)}\Vert_1={\big\Vert}\tfrac{\mathbf{x}_{i}^\top\cdot\mathbf{x}_{j}}{HW}{\big\Vert}_1={\big\Vert}\tfrac{\vert\mathbf{x}_{i}\vert\vert\mathbf{x}_{j}\vert\cos\theta}{HW}{\big\Vert}_{1},
\vspace{0.05cm}
\end{equation}
where $\mathbf{x}_{i}\in\mathbb{R}^{HW}$ denotes the $i$-th channel of the intermediate feature map $\mathbf{X}\in\mathbb{R}^{C\times HW}$. 
Note that Eq.~\eqref{eq_cov_elements_loss1} applies to the diagonal elements, and Eq.~\eqref{eq_cov_elements_loss2} applies to the off-diagonal elements of the covariance matrix.
The optimization process for the whitening loss should minimize both Eq.~\eqref{eq_cov_elements_loss1} and Eq.~\eqref{eq_cov_elements_loss2} simultaneously, but there exists a limitation on it.
The scale of each channel (\textit{i.e.,} $\vert\mathbf{x}_{i}\vert$) is forced to increase to the value of $\sqrt{HW}$ by Eq.~\eqref{eq_cov_elements_loss1} and decrease to zero by Eq.~\eqref{eq_cov_elements_loss2}. 
%while ones in off-diagonal elements will decrease to zero to satisfy Eq.~\eqref{eq_cov_elements_loss2}.%\sh{괜찮을까요? on-diagonal <> off-diagonal을 명확히하면 좋을 것 같아서 추가하였습니다! > on- /off-diagonal 문제가 아닌거 같아서 수정했습니다. 왜냐하면 전후 문장에 on/off diagonal 은 충분히 설명하고 있어서 중언인 것 같아요. be forced to 를 추가하였음}
%The scale of each channel (\textit{i.e.,} $\vert\mathbf{x}_{i}\vert$) will increase to the value of $\sqrt{HW}$ by Eq.~\eqref{eq_cov_elements_loss1} while decreasing to zero to satisfy Eq.~\eqref{eq_cov_elements_loss2}.
% In Eq.~\eqref{eq_cov_elements_loss1}, the loss attempts to increase the scale of each channel (\textit{i.e.,} $\vert\mathbf{x}_{i}\vert$) to $\sqrt{HW}$.
% However, at the same time, the same term is pushed to decrease to zero for Eq.~\eqref{eq_cov_elements_loss2}.
Therefore, forcing the diagonal and off-diagonal of the covariance matrix to be one and zero, respectively, conflicts with each other, so it is difficult to optimize both at the same time.
% \tr{이 상황이 contradiction이라고 볼수있을까요?!아니면 좀 더 가볍게 conflict?}
% \choi{conflict 인데, 실제로 optimization 이 안되구요. 간결하고 쉽게 쓰기가 어렵네요.약간 더 추가했는데 어떤가요}
% \tr{저도 조금 더 수정해봤어요!}

To address this issue, the feature map $\mathbf{X}$ can first be standardized into $\mathbf{X_s}$ through an instance normalization~\cite{ulyanov2016instance}:
\vspace{-0.1cm}
\begin{equation}\label{standardization}
% \mathbf{X_s}=\frac{\mathbf{X}-\boldsymbol{\mu}\cdot{\mathbf{1}^\top}}{\sqrt{\textit{diag}(\mathbf{\Sigma})}}
\mathbf{X_s}=\left(\text{diag}(\mathbf{\Sigma}_{\mu})\right)^{-\frac{1}{2}}\odot(\mathbf{X}-\boldsymbol{\mu}\cdot{\mathbf{1}^\top}),
\end{equation}
where $\odot$ is an element-wise multiplication, and $\text{diag}(\mathbf{\Sigma}_{\mu})\in\mathbb{R}^{C\times1}$ denotes the column vector consisting of diagonal elements in the covariance matrix. Note that each diagonal element is copied along with the spatial dimension $HW$ for element-wise multiplication.
% \begin{equation}\label{standardization}
% \mathbf{X_s}=\frac{1}{HW}\sum_{h=1}^{H}\sum_{w=1}{W}\mathbf{X}_{n}
% \end{equation}
Since the scale of each feature vector is already fixed as the unit value after the instance standardization, the whitening loss only affects the $\cos\theta$ term in Eq.~\eqref{eq_cov_elements_loss2}. In the end, this approach %\tr{loss? process? 구체적으로 this가 가리키는게 무엇인지 써주고 싶은데 적당한 표현이 어느것이 있을까요?}\choi{approach}
fits the purpose of the whitening transformation to decorrelate the features.

After standardization of the intermediate feature map, the covariance matrix is calculated as
\begin{equation}\label{eq_mean_std_covariance}
% \vspace{0.05cm}
\mathbf{\Sigma_s}=\tfrac{1}{HW}(\mathbf{X_s})(\mathbf{X_s})^\top\in\mathbb{R}^{C\times{C}},
\end{equation}
where $\mathbf{X_s}$ is the standardized feature map.
%\tr{이 문장은 빼도 좋을 것 같은데 너무 paragraph가 단순해질까요?!}
% \paragraph{Instance whitening loss (Fig.~\ref{fig:whitening_overview}(c, d))}
Thanks to the standardization process, diagonal elements of the covariance matrix are already set as unit values. Thus, we only need to make the off-diagonals of the covariance matrix close to zero, which makes it easy to optimize for the whitening process, and the aforementioned conflict can thus be resolved. Since the covariance matrix is symmetric, the loss can be applied only to the strict upper triangular part. Our instance whitening (IW) loss is formulated as
\begin{equation} \label{eq_our_whitening_loss}
\mathcal{L}_{\text{IW}} = \mathbb{E}[\Vert\mathbf{\Sigma_s}\odot\mathbf{M}\Vert_1],
\end{equation}
where $\mathbb{E}$ denotes the arithmetic mean and $\mathbf{M}\in\mathbb{R}^{C\times C}$ denotes a strict upper triangular matrix, \textit{i.e.,}
\begin{equation} \label{eq_mask}
\mathbf{M}_{i,j} =\begin{cases}
 0, & \text{if $i \geq j$}\\
 1, & \text{otherwise}
\end{cases} \quad
0\le i,j<C.
\vspace*{-0.1cm}
\end{equation}
% \tr{vspace로 지치면 빼도 좋을 것 같은 eq}

\begin{figure*}[ht!]
\vspace*{-0.0cm}
  \centering\includegraphics[width=0.97\linewidth]{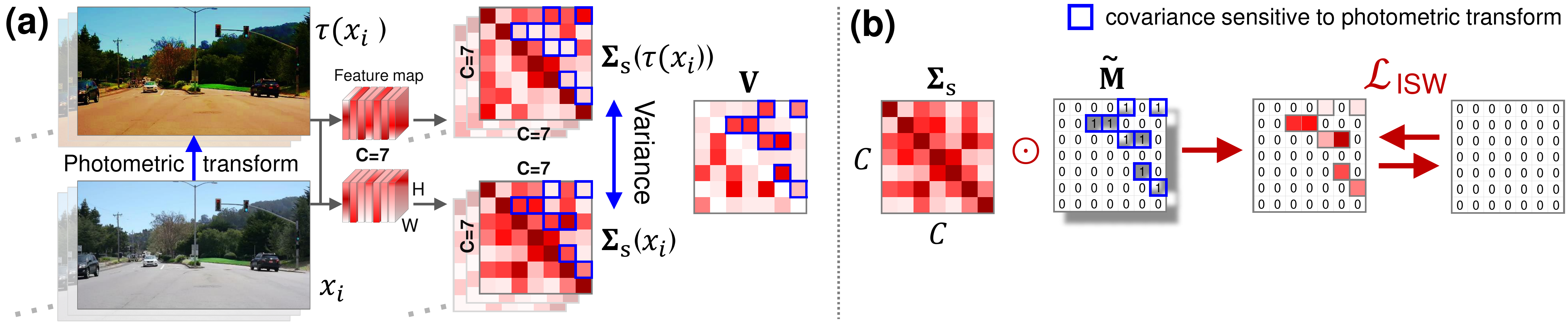}
  \vspace*{-0.0cm}
  \caption{\textbf{Instance selective whitening loss}. 
  (a) 
  The variance matrix $\mathbf{V}$ is computed out of the covariance matrices of the $i$-th image $x_i$ and its photometric transformed image $\tau(x_i)$ to identify those elements sensitive to the transformation (blue boxes).
  Note that these matrices are symmetric.
  (b) The covariance matrix $\mathbf{\Sigma_s}$ is masked by the matrix $\mathbf{\tilde{M}}$ to selectively suppress style-sensitive covariances by $\mathcal{L}_{\text{ISW}}$.
%   Notations; $x_i$: $i$-th image, $\tau(x_i)$: photometric transformed image of $x_i$, $\mathbf{\Sigma_s}(x_i)$: covariance matrix calculated from the feature map of $x_i$, $\mathbf{V}$: the matrix consists of elements of the variance values of each covariance elements, $\mathcal{L}_{\text{IW}}$: our proposed instance selective whitening loss. \choi{caption 봐주세요}
  %   Matrix (bold) notations stand for feature maps;
  %   \todo{Feature map notations;}
%   Each operation $op$ is notated as $\text{G}_{op}$, and feature maps are in bold--$\text{X}_{\ell}$: lower-level feature map, Z: width-wise pooled $\text{X}_{\ell}$, $\hat{\text{Z}}$: down-sampled Z, $\text{Q}^{n}$: $n$-th intermediate feature map of 1D convolution layers, $\hat{\text{A}}$: down-sampled attention map, A: final attention map, $\text{X}_{h}$: higher-level feature map, $\Tilde{\text{X}}_{h}$: transformed new feature map. Details can be found in Section~\ref{sec:method_block}.
}
\label{fig:disentangling_covariance}
\vspace{-0.45cm}
\end{figure*}

\vspace*{-0.18cm}
\subsection{Margin-based relaxation of whitening loss}
\vspace*{-0.12cm}
The instance whitening loss (Eq.~\eqref{eq_our_whitening_loss}) suppresses all covariance elements to zero, so it can adversely affect the discriminative power of features within DNNs. To address this issue, we propose an instance-\emph{relaxed} whitening (IRW) loss %as a hinge loss variant
to sustain the covariance elements essential in maintaining the discriminative power. The IRW loss is designed so that the expected value of the total covariance lies within a specified margin $\delta$ rather than being close to zero, \textit{i.e.,} 
\begin{equation} \label{eq_our_whitening_loss_with_margin}
\mathcal{L}_{\text{IRW}} = \max(\mathbb{E}[\Vert\mathbf{\Sigma_s}\odot\mathbf{M}\Vert_1]-\delta, \ 0)
\end{equation}
The loss $\mathcal{L}_{\text{IRW}}$ allows the covariance to have a certain level of values, so it gives room to keep discriminative features intact. The empirical effect of the IRW loss can be found in Section~\ref{exp:effectiveness}. It shows better performance compared to the IW loss not including margin $\delta$ (Eq.~\eqref{eq_our_whitening_loss}). Nonetheless, it may not be sufficient because we cannot guarantee that only the covariance useful for generalization performance remains through the margin relaxation.

\vspace*{-0.05cm}
\subsection{Separating Covariance Elements}\label{method:deep_instance_selective_whitening}
\vspace*{-0.1cm}
% \out{Covariance 의 Disentangling 을 해야함. Covariance가 인코딩한 정보 중, domain gap에 영향을 주는 covariance들을 선별적으로 없애는 것이 intuition이고, 우리는 color, style, camera distortion (blurry)등이 domain gap에 많은 부분을 차지하는걸 알고있으므로 이들을 선별적으로 제거하기위해 instance whitening을 적용하지 않은 모델의 input image, color augmented image, blurred image의 intermediate feature map을 구하여 명시적으로 covariance의 차이가 크게 나는 부분들을 선별하여 이를 고정적으로 training시에 없앤다. top n 개를 뽑기위해 우리는 kmeans clustering, k=2를 주어 covariance가 많이 차이나는 경계를 구했다.} \out{Color augmentation, Blur augmentation 을 통해서 style 에 해당하는 factor 에 영향을 받는 Covariance 를 구함} \newline
% \sh{이전 섹션의 문제제기부터 시작해서 마지막까지 논리 되게 좋은 것 같아요!}
To further improve our approach, we need to separate the covariance terms into two groups: domain-specific style and domain-invariant content. We propose to selectively suppress only the style-encoded covariances that cause the domain shift. Assuming that the domain shift includes changes in color and blurriness, we simulate the domain shift through photometric augmentation such as color jittering and Gaussian blurring.

First, we add only the instance standardization layer into the networks (Fig.~\ref{fig:whitening_overview}(a)) and train them during the $n$ initial epochs
%\footnote{Note that $n$ is a hyperparameter, but it does not affect performance sensitively. See the supplementary section.}
without the whitening loss to get the pure statistics of the covariance matrices from training images. $n$ is a hyper-parameter, which we empirically set to 5. Afterwards, we extract two covariance matrices by inferring from two input images, namely an original and a photometric-transformed image, and calculate the variance matrix from the differences between two different covariance matrices. 
%\yun{IS를 적용하여 train후, high variance를 가지는 그룹에 ISW를 적용하여 학습을 다시 진행한다는 부분 명시적으로 적으면 어떨까요? infer하고 loss 제안한다로 끝나버려서요..}
% At the next step, we inference two images which are the original and photometrically augmented images, and calculate the covariance difference between them. We assume the differences of the covariances between them stands for the sensitivity to the color or blurriness of images, which means the corresponding covariance elements encode the photometric information.
% \mathbf{\Sigma}_{\mu}=\tfrac{1}{HW}\left(\mathbf{X}-\boldsymbol{\mu}\cdot{\mathbf{1}^\top}\right)\left(\mathbf{X}-\boldsymbol{\mu}\cdot{\mathbf{1}^\top}\right)^\top\in\mathbb{R}^{C\times{C}}.
%^{-\frac{1}{2}}\left(\mathbf{X}-\boldsymbol{\mu}\cdot{\mathbf{1}^\top}\right)
%consists of elements of the variance for each covariance element variances for each covariance is
Formally, the variance matrix $\mathbf{V}\in\mathbb{R}^{C\times C}$ is defined as
\vspace*{-0.1cm}
\begin{equation} \label{eq_var_covar_sigma}
\mathbf{V}=\frac{1}{N}\sum_{i=1}^N\boldsymbol{\sigma}_i^2,
\end{equation}
from mean $\boldsymbol{\mu}_{\mathbf{\Sigma}_i}$ and variance $\boldsymbol{\sigma}_i^2$ for each element from two different covariance matrices of the $i$-th image, \textit{i.e.,}
\vspace*{-0.1cm}
\begin{equation} \label{eq_var_covar_mu}
\boldsymbol{\mu}_{\mathbf{\Sigma}_i}=\tfrac{1}{2}\left(\mathbf{\Sigma_s}({x_i})+ \mathbf{\Sigma_s}(\tau(x_i)\right)
\vspace*{-0.2cm}
\end{equation}
\begin{equation} \label{eq_var_covar_sig}
\boldsymbol{\sigma}_i^2=\tfrac{1}{2}\left(\left(\mathbf{\Sigma_s}\left({x_i}\right)-\boldsymbol{\mu}_{\mathbf{\Sigma}_i}\right)^2 + \left(\mathbf{\Sigma_s}\left(\tau\left(x_i\right)\right)-\boldsymbol{\mu}_{\mathbf{\Sigma}_i}\right)^2\right)
\vspace*{-0.1cm}
\end{equation}
% \sh{혹시 eq.12 에서 한 식으로 표현하는 것 대신 두 Cov matrix의 차를 구한후에 거기에 분산을 넣는건 어떨까요? 저 분산 연산은 한개의 인풋만 받을 수 있을것 같은 느낌이 들어서요ㅠ}\yun{eq.12가 N개의 이미지 페어에 대해 covariance의 variance를 구한다는 의미죠? N에 대한 설명을 적어주면 좋을것같아요}
where $N$ denotes the number of image samples, $x_i$ is the $i$-th image sample, $\tau$ is a photometric transformation, and $\mathbf{\Sigma_s}(\cdot)$ extracts the covariance matrix of the intermediate feature map from an input image.
As a result, $\mathbf{V}$ consists of elements of the variance of each covariance element across various photometric transformations.
% \yun{이 마지막 문장, covariance의 variance다라고 쓰면 더 간단하지 않을까요?}

We assume that the variance matrix $\textbf{V}$ implies the sensitivity of the corresponding covariance to the photometric transformation. This means that the covariance elements with high variance value contain the domain-specific style such as color and blurriness.
%\sh{Covariance encode 표현 --> contain으로 변경}
To identify such elements, we apply \textit{k}-means clustering on the strict upper triangular elements $\mathbf{V}_{i,j}\; (i<j)$ of the variance matrix $\mathbf{V}$ to assign the elements into $k$ clusters $C=\{c_1, c_2,\dots,c_k\}$ with respect to the value. 
%\sh{혹시 여기서 increasing order라고 써주는게 도움이 될까요?}
Next, we split the $k$ clusters into two groups, $G_{low}=\{c_1,\dots,c_m\}$ with low variance value and $G_{high}=\{c_{m+1},\dots,c_k\}$ with high variance value.
% Next, we pick a cluster which has the smallest variance values and treat it as a set of covariances that contains the domain-invariant content information. On the other hand, we assume the others contain domain-specific style information. Note that $k$ is a hyper-parameter and is empirically set to 50. More details can be found in the supplementary section.
The hyper-parameters $k$ and $m$ are empirically set to 3 and 1, respectively. More details can be found in the supplementary Section \todow{A.2}.
% \sh{$m$ 제거 후에 k cluster에 대한 top 1을 뽑도록 설명 변경함, 확인부탁드려요...} \choi{뒤에 G_{high} 가 수식에 나오기 때문에 그냥 원래대로 할께요 :)}\sh{넵!} 
%We assume that the style information is encoded in the covariances belonging to $G_{high}$ and content information is encoded in the covariances belonging to $G_{low}$.
We assume that $G_{high}$ contains the domain-specific style and $G_{low}$ contains domain-invariant content.
%\sh{Ditto. encoding wording > encoding, contain 둘 다 그냥 갑시다}
% \yun{'s'가 standardization과 style을 의미하는게 섞이는데 저는 좀 헷갈리네요.ㅠ}
%\tr{$G_{h}$ 이거 자주 안나올 예정이면 $G_\text{high}, G_{low}$로 명시하면 가독성에 좋을 것 같은데 어떠신가요} 

Finally, we propose an instance \emph{selective} whitening (ISW) loss that selectively suppresses only to the style-encoded covariances.
Let the mask matrix $\mathbf{{M}}$ in Eq.~\eqref{eq_mask} change to $\mathbf{\tilde{M}}\in\mathbb{R}^{C \times C}$ for the ISW loss as
% The matrix $\mathbf{\tilde{M}}\in\mathbb{R}^{C \times C}$ for leaving only the covariance to which the whitening loss is applied is defined as follows.
\vspace{-0.1cm}
\begin{equation} \label{eq_selective_mask}
\mathbf{\tilde{M}}_{i,j} =\begin{cases}
 1, & \text{if $\mathbf{V}_{i,j}\in G_{high}$}\\
 0, & \text{otherwise}
\end{cases} \quad
\vspace*{-0.05cm}
\end{equation}
The ISW loss is defined as
\vspace*{-0.05cm}
\begin{equation} \label{eq_our_whitening_loss_selective}
\mathcal{L}_{\text{ISW}} = \mathbb{E}[\Vert\mathbf{\Sigma_s}\odot\mathbf{\tilde{M}}\Vert_1].
\vspace*{-0.05cm}
\end{equation}
The networks continue training for the remaining epochs incorporating the proposed ISW loss.

\begin{figure}[b!]
\vspace{-0.3cm}
\begin{center}
  \includegraphics[width=1.0\linewidth]{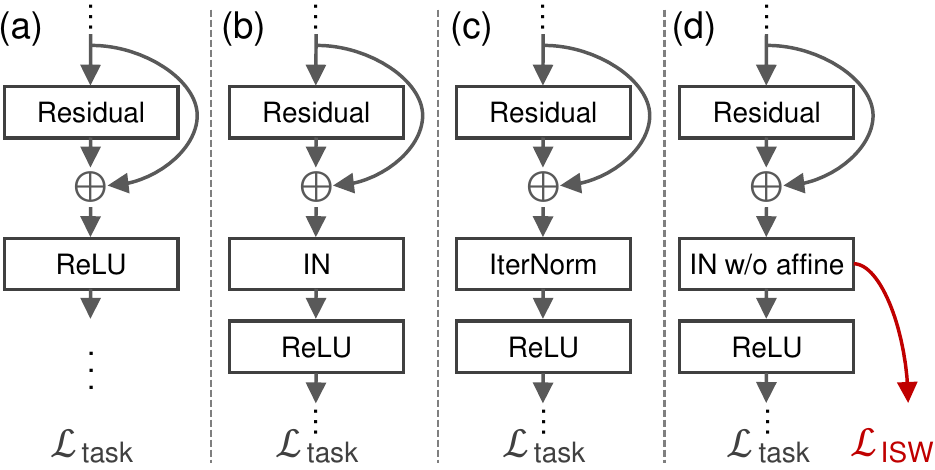}
\end{center}
\vspace*{-0.3cm}
   \caption{\textbf{Architecture comparison with other methods}: (a) the original residual block~\cite{he2016deep}; (b) IBN-b~\cite{pan2018two} combining instance normalization with batch normalization; (c) IterNorm~\cite{huang2019iterative} employing Newton's iterations for efficient whitening; (d) Our proposed ISW loss applied to instance normalization.
%   For fair comparison, we reimplement (b) IBN-b and (c) IterNorm on our baseline.
   }
%   the order of the channels are changed by clustering.}
%   Y 축은 height 이고 X축은 채널입니다. 즉 각 vertical position 마다 어떤 channel 을 attention 하고 있는지를 나타냄. Note that 여기서의 채널 순서는 보다 나은 보기를 위해서, 비슷한 채널 들끼리 clustering 하였습니다.}
\label{fig:architectural_comparison}
\end{figure}

\vspace*{-0.0cm}
\subsection{Network architecture with proposed ISW loss}\label{sec:network_architecture}
\vspace*{-0.05cm}
% \vspace{-0.1cm}
IBN-Net~\cite{pan2018two} has explored a number of ResNet~\cite{he2016deep}-based architectures to combine instance normalization with batch normalization and proposed several IBN blocks based on a residual block (Fig.~\ref{fig:architectural_comparison}(a)). Among the proposed blocks, IBN-b, which adds an instance normalization layer right after the addition operation of a residual block (Fig.~\ref{fig:architectural_comparison}(b)), shows the best generalization performance on semantic segmentation tasks. After all, they add three instance normalization layers after the first three convolution groups (\textit{i.e.,} conv1, conv2\_x, and conv3\_x). We follow this architectural approach as our baseline. As shown in Fig.~\ref{fig:architectural_comparison}(d), we simply add our proposed ISW loss to the instance normalization layer.
% \sh{@shchoi 여기 추가했습니다}For the first convolution layer, we have modified architecture of three 3$\times$3 convolution layers. Hence, we add our proposed ISW loss to the last layer of three.
Our loss in total is described as
\vspace{-0.25cm}
\begin{equation} \label{eq_final_loss}
\mathcal{L}_{\text{total}} = \mathcal{L}_{\text{task}} + \lambda(\frac{1}{L}\sum_i^L\mathcal{L}_{\text{ISW}}^i),
\vspace{-0.15cm}
\end{equation}
where $\lambda$ denotes the weight of our ISW loss and is empirically set to 0.6, $\mathcal{L}_{\text{task}}$ is the task loss (\textit{e.g.,} a per-pixel cross-entropy loss for semantic segmentation), $i$ indicates the layer index, and $L$ is the number of layers to which the ISW loss is applied. The hyper-parameter $\lambda$ is analyzed in the supplementary Section \todow{A.2}. $L$ is set to three by following IBN-Net.
An affine transformation is not used since the subsequent convolution operation after a whitening transformation can do the equivalent job, and empirically, we found no performance gain by explicitly adding the affine transformation.
\vspace{-0.1cm}

\vspace{-0.0cm}
\section{Experiments}
\vspace{-0.1cm}
This section describes the experimental setup and presents evaluation results to assess the effectiveness of our proposed methods on semantic segmentation with comparison to other methods. Furthermore, we provide an in-depth analysis of our results including the covariance matrices.
\vspace{-0.07cm}
\subsection{Experimental Setup}
\vspace{-0.08cm}
We train our model on several datasets (\textit{e.g.,} Cityscapes) and show its performance on other datasets (\textit{e.g.,} BDD-100K, Mapillary, GTAV, and SYNTHIA) to measure the generalization capability on unseen domains.
% There exists only one DG work, which suggests the normalization technique for semantic segmentation~\cite{pan2018two}.\choi{확실하지 않을 수 있는데, 단정된 주장이라고 생각합니다}
For fair comparisons with other normalization techniques, we re-implement IBN-Net~\cite{pan2018two} and IterNorm~\cite{huang2019iterative} on our baseline models and compare them with our methods.
%\sh{Fair comparison 구문 전부 확인해볼것 -> a fair comparison 혹은 fair comparisons로 변경}
%by ourselves. 
As described in Section~\ref{sec:network_architecture}, our proposed loss can easily be added to existing models, so we apply our methods to various backbone networks such as 
% As we pointed out, our work can be deployed with any DA or DG approaches,
% as our approach proposes an auxiliary loss without any modification of the architecture.\sh{CHECK - auxiliary loss}\choi{any modification 은 아닌 것 같아요}
% To demonstrate this aspect, we apply the instance selective whitening loss to various backbone networks such as
ResNet~\cite{he2016deep}, ShuffleNetV2~\cite{ma2018shufflenet} and MobileNetV2~\cite{sandler2018mobilenetv2} and show wide applicability of the proposed methods.
For all the quantitative experiments, mean Intersection over Union (mIoU) is used to measure the segmentation performance.
\vspace{-0.4cm}
\subsubsection{Implementation details}
\vspace{-0.1cm}
%\tr{이 paragraph (dataset도?) 5.1로 옮겨도 좋을 것 같은데 어떻게 생각하시나요. 5.3에서 보여줄 결과가 따로 실험세팅이 많이 달라지나요?!}\sh{원래는 classification 실험도 싣을 예정이었어서 5.2 안에 넣었네요, 밖으로 뺄게요 :)}
We adopt DeepLabV3+~\cite{chen2018encoder} for a semantic segmentation architecture, and SGD optimizer with an initial learning rate of 1e-2 and momentum of 0.9 is used. Besides, we follow the polynomial learning rate scheduling~\cite{liu2015parsenet} with the power of 0.9. We train all the models for 40K iteration, except for multi-source models, which are trained for 110K iterations. To prevent the model from overfitting, color and positional augmentations such as color jittering, Gaussian blur, random cropping, random horizontal flipping, and random scaling with the range of [0.5, 2.0] are conducted. For the photometric transformation in ISW, we apply color jittering and Gaussian blur.
% Note that style information tends to reside in the feature maps from early layers, as IBN-net~\cite{pan2018two} also pointed out. Hence, we introduce the instance whitening at the end of the first three convolution groups.
Also, as suggested by IBN-Net, we add three instance normalization layers after the first three convolution groups and apply our proposed loss.
% We perform an instance standardization over the input images as one of the image pre-processing steps.
% Instance whitening loss applies after the certain number of epochs are done to obtain the pure statistics.\sh{CHECK - pure statistics}
% In our experiments, we calculate the statistics after 5 and 10 epochs for the single-source and multi-source experiments, respectively. We set the instance whitening loss weight to 0.8.
Further details are provided in the supplementary Section \todow{A.3}.
\vspace{-0.35cm}
\subsubsection{Datasets}
\vspace{-0.2cm}
To verify the generalization capability of our methods, we conduct the experiments on five different datasets.
\vspace{-0.48cm}
\paragraph{Real-world datasets} Cityscapes~\cite{Cordts2016Cityscapes} is a large-scale dataset containing high-resolution (\textit{e.g.,} 2048$\times$1024) urban scene images collected from 50 different cities in primarily Germany. It provides 3,450 finely-annotated images and 20,000 coarsely-annotated images. We use only a finely-annotated set for training and validation.
%\tr{이 fine annotation set이 바로 직전에 말한 finely-annotated images랑 구성이 다른가요!?}\sh{같아요, 중복 삭제했어요}
BDD-100K~\cite{yu2020bdd100k} is another real-world dataset that contains diverse urban driving scene images %, randomly sampled from 100K driving videos 
with the resolution of 1280$\times$720. The images are collected from various locations in the US. For a semantic segmentation task, 7,000 training and 1,000 validation images are provided.
The last real-world dataset we use is Mapillary~\cite{neuhold2017mapillary}, a diverse street-view dataset consisting of 25,000 high-resolution images with a minimum resolution of 1920$\times$1080 collected from all around the world.
%It provides 66 object labels, including 19 objects of Cityscapes.

\vspace{-0.48cm}
\paragraph{Synthetic datasets} GTAV~\cite{Richter_2016_ECCV} is a large-scale dataset containing 24,966 driving-scene images generated from Grand Theft Auto V game engine. It has 12,403, 6,382, and 6,181 images of size 1914$\times$1052 for a train, a validation, and a test set, respectively. It has 19 object categories compatible with Cityscapes.
Also, we use SYNTHIA~\cite{ros2016synthia}, composed of photo-realistic synthetic images containing 9,400 samples with a resolution of 960$\times$720. 
% \vspace{-0.2cm}

\vspace{-0.1cm}
\subsection{Quantitative Evaluation}
\vspace{-0.1cm}
This subsection provides ablation studies, the comparisons of our results against other normalization methods, the evaluation on multiple source domains, and the analysis of computational cost. Since the experiments follow domain generalization settings, the model cannot access any datasets other than the source data.
% We start by explaining the effectiveness of instance selective whitening loss, and then present the comparison with previous approaches~\cite{pan2018two, yue2019domain}. 
\vspace{-0.38cm}
\subsubsection{Effectiveness of instance selective whitening loss}\label{exp:effectiveness}
\vspace{-0.17cm}
% To show the effect of ISW loss, we conduct extensive experiments.
To verify the effectiveness of our methods, we conduct comparisons with other normalization methods and ablation studies on instance whitening (IW), instance-relaxed whitening (IRW), and instance selective whitening (ISW).
\textbf{Note that all the experiments in this subsection are performed three times and averaged for fair comparisons}. % For the fair comparison, we implement $^\dagger$IBN-Net~\cite{pan2018two} and baselines with instance whitening.
%To explain selectively suppressing the covariance enhances the generalization performance, 
% we compare our final model with instance whitening methods and a model with IRW loss.
%To explain our performance gain doesn't come from adopting the instance whitening, but from selectively suppressing the covariances, we run baselines with instance whitening methods and IRW loss.

\begin{table}[t!]
\vspace*{-0.0cm}
\begin{center}
\footnotesize
\begin{tabular}{c|c|c|c|c||c}
\toprule
Models (GTAV) & C & B & M & S & G\\
% Models & G$\rightarrow$ C & G$\rightarrow$ B & G$\rightarrow$ M & G$\rightarrow$ G & G$\rightarrow$ S \\
\drule
Baseline  & 28.95        & 25.14      & 28.18      & 26.23      & 73.45 \\ 
\midrule
$^\dagger$SW~\cite{pan2019switchable} & 29.91      & 27.48      & 29.71      & 27.61      & \textbf{73.50}     \\ 
\midrule
$^\dagger$IBN-Net~\cite{pan2018two}      & 33.85      & 32.30      & 37.75      & 27.90 & 72.90      \\ 
\midrule
$^\dagger$IterNorm~\cite{huang2019iterative} & 31.81      & 32.70      & 33.88      & 27.07      & 73.19     \\ 
\midrule
Ours (IW) & 33.21      & 32.67      & 37.35      & 27.57      & 72.06      \\ 
\midrule
Ours (IRW)                   & 33.57      & 33.18      & 38.42      & 27.29      & 71.96      \\ 
\midrule
Ours (ISW)                   & \textbf{36.58}      & \textbf{35.20} & \textbf{40.33} & \textbf{28.30}      & 72.10      \\ 
\bottomrule
\end{tabular}
\end{center}
\vspace*{-0.15cm}
\caption{Comparison of mIoU(\%). Compared models are trained on GTAV train set, and validated on Cityscapes (C), BDD-100K (B), Mapillary (M), SYNTHIA (S) and GTAV (G) validation sets. ResNet-50 with an output stride of 16 is used.
%ResNet-50 and output stride of 16 on Cityscapes (C), BDD-100K (B), Mapillary (M), SYNTHIA (S), and GTAV (G) validation sets.
$^\dagger$ denotes our own re-implemented models. SW denotes Switchable Whitening~\cite{pan2019switchable}.}
\label{tab_best_model_16_gtav}
\vspace*{-0.53cm}
\end{table}

Table~\ref{tab_best_model_16_gtav} shows the generalization performance of the models trained on GTAV dataset. ISW outperforms other methods on all datasets except the source dataset (\textit{i.e.,} GTAV). Especially, ISW shows a significant improvement on real-world datasets (i.e., Cityscapes, BDD-100K, and Mapillary). Table~\ref{tab_best_model_16_city} shows the generalization performance of those models trained on Cityscapes dataset. 
Although IterNorm outperforms our models on GTAV, the performance gap is minimal. 
ISW outperforms other normalization and baseline models on BDD-100K, Mapillary, and SYNTHIA datasets. % with a large margin.}
%All of our methods outperform other methods on real-world datasets. Especially, on BDD-100K and Mapillary datasets, our methods show a large performance margin compared to other methods, and ISW shows significant improvement.

\begin{table}[b!]
\vspace{-0.4cm}
\begin{center}
\footnotesize
\begin{tabular}{c|c|c|c|c||c}
\toprule
Models (Cityscapes) & B & M & G & S & C \\
\drule
Baseline  & 44.96      & 51.68      & 42.55      & 23.29      & \textbf{77.51}  \\ 
\midrule
$^\dagger$SW~\cite{pan2019switchable} & 48.49      & 55.82 & 44.87      & 26.10      & 77.30       \\ 
\midrule
$^\dagger$IBN-Net~\cite{pan2018two}      & 48.56      & 57.04      & 45.06      & 26.14      & 76.55      \\ 
\midrule
$^\dagger$IterNorm~\cite{huang2019iterative} & 49.23      & 56.26 & \textbf{45.73}      & 25.98      & 76.02       \\ 
\midrule
Ours (IW) & 48.19      & 58.90      & 45.21      & 25.81      & 76.06      \\ 
\midrule
Ours (IRW)                   & 48.67 & \textbf{59.20} & 45.64
& 26.05      & 76.13      \\ 
\midrule
Ours (ISW)                   & \textbf{50.73} & 58.64 & 45.00 & \textbf{26.20} & 76.41      \\ 
\bottomrule
\end{tabular}
\end{center}
\vspace*{-0.17cm}
\caption{Comparison of mIoU(\%). The models are trained on Cityscapes train set. ResNet-50 with an output stride of 16 is used.
%ResNet-50 and output stride 16 on five different validation sets.
$^\dagger$ denotes re-implemented models.}
\label{tab_best_model_16_city}
\vspace{-0.2cm}
\end{table}

Baseline, Switchable Whitening (SW), and IBN-Net, which are less generalizable than our method, tend to overfit the source domain, suffering from performance degradation on the target domain due to the large domain shift. 
% In contrast, our method may sacrifice the minimal amount of performance on the source domains (\textit{i.e.,} training and evaluating on the same dataset). However, when deployed in the wild, we believe that the performance on the real-world data with large domain-shift is more critical.
% Our method sacrifices the minimal amount of performance on the source domains (\textit{i.e.,} training and evaluating on the same dataset). 
Our method may sacrifice the performance on the source domains (\textit{i.e.,} training and evaluating on the same dataset) as shown in the last column in Table~\ref{tab_best_model_16_gtav} and ~\ref{tab_best_model_16_city}. 
However, our models shows good generalizability, which is critical when deployed in the wild, where large domain-shift is expected.

% Note that the validation results of ISW on the source domains (\textit{i.e.,} training and evaluating on the same dataset) show better performance to IW and IRW as shown in the last column in Table~\ref{tab_best_model_16_gtav} and ~\ref{tab_best_model_16_city}. This comes from the fact that ISW suppresses the style-sensitive covariances only, while IW and IRW removes all of them.

Table~\ref{tab_various_backbone_cityscapes} explains the wide applicability of our work. The first group is reported by adopting ShuffleNetV2, and the second group is using MobileNetV2 as backbone networks. In both cases, our model with ISW outperforms the baseline and IBN-Net on real-world datasets.
% Baseline and IBN-Net, which are less generalizable than our method, tend to overfit to the source domain, 
% %it has good performance on the same domain (e.g. GTAV$\rightarrow$GTAV), but
% suffering from performance degradation due to large domain shift. 
% In contrast, our method may sacrifice the minimal amount of performance if the domain gap is small (\textit{e.g.,} the same synthetic datasets: G$\rightarrow$S, G). 
% However, when deployed in the wild, we believe that the performance on the real-world data (\textit{e.g.,} G$\rightarrow$C, S, M; C$\rightarrow$B, M) with large domain-shift is more critical.
% Note that the validation results of ISW on the source domains (\textit{i.e.,} training and evaluating on the same dataset) show comparable performance to IBN-Net as shown in the last column in Table~\ref{tab_best_model_16_gtav} and ~\ref{tab_best_model_16_city}. In contrast, there are significant performance drops on other whitening methods such as IterNorm and IW. This comes from the fact that ISW suppresses the style-sensitive covariances only, while instance whitening removes all of them. Table~\ref{tab_various_backbone_cityscapes} explains the wide applicability of our work. The first group is reported by adopting ShuffleNetV2, and the second group is using MobileNetV2 as backbone networks. In both cases, our model with ISW outperforms the baseline and IBN-Net.
% \choi{Table 1 과 2에서 ISW 가 IW 나 IRW 를 이긴다는 거 언급 해주세요. 모든 covariance 를 suppress 하지 않고, style sensitive 한 covariance 만 선별적으로 제거해서..또한 covariance 를 없애지 않는 IBNNet 과도 비슷하다.} \sh{Check 부탁드려요~}
To further validate the capability of our method, we present the comparison with baselines trained on multiple synthetic domains, GTAV, and SYNTHIA. For the training, we aggregate the training domains without any joint training methodologies.
Learning domain-invariant features across multiple datasets is essential to optimize the model on different distributions of multiple datasets. Table~\ref{tab_multi_source_best} shows our model trained on multiple datasets performs better than other models due to its generalization ability by extracting domain-invariant features during training.
% Table~\ref{tab_multi_source_best} shows our method with ISW outperforms the baseline and IBN-Net~\cite{pan2018two} with large improvements for all three domains.
% Note that there is a significant domain gap between the real and synthetic dataset. This result shows our model has substantial generalization capability. 

\begin{table}[t!]
\vspace*{-0.0cm}
\begin{center}
\setlength\tabcolsep{5.2pt}
\footnotesize
\begin{tabular}{c|c|c|c|c||c}
\toprule
Models (GTAV) & C & B & M & S & G \\
\drule
% Baseline              & -37.62      & 44.06      & 35.56      & 21.11      & \textbf{72.89} \\
Baseline              & 25.56      & 22.17      & 28.60      & 23.33      & \textbf{66.47} \\
\midrule
% $^\dagger$IBN-Net~\cite{pan2018two} & -40.92      & \textbf{49.91} & 38.07      & 22.28      & 71.16      \\
$^\dagger$IBN-Net~\cite{pan2018two} & 27.10     & 31.82 & 34.89      & \textbf{25.56}      & 65.44      \\
\midrule
% Ours (ISW)                & \textbf{-42.70} & 49.60      & \textbf{40.81} & \textbf{22.43} & 71.03      \\
Ours (ISW)                & \textbf{30.98} & \textbf{32.06}      & \textbf{35.31} & 24.31 & 64.99      \\
\toprule
Baseline              & 25.92      & 25.73      & 26.45  & 24.03      & \textbf{68.12} \\
\midrule
$^\dagger$IBN-Net~\cite{pan2018two} & 30.14      & 27.66      & 27.07      & \textbf{24.98}      & 67.66      \\
\midrule
Ours (ISW)              & \textbf{30.86} & \textbf{30.05} & \textbf{30.67} & 24.43 & 67.48      \\
\bottomrule
\end{tabular}
\end{center}
\vspace*{-0.15cm}
\caption{Comparison of mIoU(\%). The models are trained on GTAV train set. The backbone networks of the first group are ShuffleNetV2~\cite{ma2018shufflenet} and the second group is MobileNetV2~\cite{sandler2018mobilenetv2}.}%$^\dagger$ denotes re-implemented models.}
\label{tab_various_backbone_cityscapes}
\vspace*{-0.55cm}
\end{table}

\begin{table}[h]
\vspace*{-0.15cm}
% \hspace*{-0.2cm}
\begin{center}
\footnotesize
\begin{tabular}{c|c|c|c||c|c}
\toprule
Models (G + S) & C & B & M & G & S \\
\drule
Baseline & 35.46 & 25.09 & 31.94 & 68.48 & 67.99 \\ 
% Baseline & -33.16 & 24.17 & 31.39 & 65.90 & 69.29 \\ 
\midrule
IBN-Net & 35.55 & 32.18 & 38.09 & \textbf{69.72} & 66.90 \\
% IBN-Net & -33.47 & 29.79 & 34.46 & \textbf{68.75} & 70.04 \\
\midrule
\textbf{Ours} & \textbf{37.69} & \textbf{34.09} & \textbf{38.49} & 68.26 & \textbf{68.77} \\
% \midrule
% cluster 5 / weight 0.6
% \textbf{Ours\_temp} & \textbf{36.91} & \textbf{34.45} & \textbf{38.52} & 68.68 & \textbf{68.61} \\
% \textbf{Ours} & \textbf{-36.64} & \textbf{32.19} & \textbf{36.29}  & 66.95 & \textbf{73.42}\\
% \begin{tabular}{c|c|c|c}
% \toprule
% Models (G + S) & C & B & M \\
% \drule
% Baseline              & 34.51      & 23.30      & 29.11      \\ 
% %Baseline    (previous)       & 33.16      & 24.17      & 31.39      \\ 
% \midrule
% $^\dagger$IBN-Net~\cite{pan2018two} & 33.47      & 29.79      & 34.46      \\ 
% \midrule
% \textbf{Ours}             & \textbf{36.64} & \textbf{32.19} & \textbf{36.29} \\
\bottomrule
\end{tabular}
\end{center}
\vspace*{-0.1cm}
\caption{Comparison of mIoU(\%). The models are trained on multiple synthetic domains. The backbone is ResNet-50 with an output stride of 16. $^\dagger$ denotes re-implemented models.}
\label{tab_multi_source_best}
\vspace*{-0.5cm}
\end{table}

% \begin{table}[ht!]
% \vspace*{-0.2cm}
% \begin{center}
% \setlength\tabcolsep{4.5pt}
% \footnotesize
% \begin{tabular}{c|c|c||c|c|c|c}
% \toprule
% Backbones & Models & C & B & M & G & S \\
% \drule
% \multirow{3}{*}{\shortstack{ShuffleNet\\V2~\cite{ma2018shufflenet}}} & Baseline & \textbf{72.89} & 37.62 & 44.06 & 35.56 & 21.11 \\
% \cmidrule{2-7}
% & $^\dagger$IBN-Net~\cite{pan2018two} & 71.16 & 40.92 & \textbf{49.91} & 38.07 & 22.28 \\
% \cmidrule{2-7}
% & Ours & 71.03 & \textbf{42.70} & 49.60 & \textbf{40.81} & \textbf{22.43} \\
% \midrule
% \multirow{3}{*}{\shortstack{MoblieNet\\V2~\cite{sandler2018mobilenetv2}}} & Baseline & \textbf{75.69} & 40.45 & 50.17 & 36.06 & 20.43 \\
% \cmidrule{2-7}
% & $^\dagger$IBN-Net~\cite{pan2018two} & 75.01 & 43.97 & 50.28 & 40.13 & 22.02 \\
% \cmidrule{2-7}
% & Ours & 74.78& \textbf{45.62} & \textbf{52.94} & \textbf{42.55} & \textbf{23.57} \\
% \bottomrule
% \end{tabular}
% \end{center}
% \vspace*{-0.1cm}
% \caption{mIoU(\%) comparison between various backbone networks trained on Cityscapes dataset. Backbone architecture is ResNet-50 and output stride 16. $^\dagger$ denotes the model implemented by authors.}
% \label{tab_various_backbone_cityscapes}
% \vspace*{-0.3cm}
% \end{table}

\begin{table}[b!]
\vspace*{-0.0cm}
\begin{center}
\setlength\tabcolsep{4.2pt}
\footnotesize
\begin{tabular}{c|cc|cc|cc}
\toprule
Models (GTAV) & \multicolumn{2}{c|}{C} & \multicolumn{2}{c|}{B} & \multicolumn{2}{c}{M} \\
\drule
Baseline & 22.20   & \multirow{2}{*}{7.40 $\uparrow$} & \multicolumn{2}{c|}{\multirow{2}{*}{N/A}}  & \multicolumn{2}{c}{\multirow{2}{*}{N/A}} \\ 
IBN-Net~\cite{pan2018two} & 29.60   & & & &  & \\ 
\midrule
Baseline          & 32.45 & \multirow{2}{*}{4.97$\uparrow$} & 26.73 & \multirow{2}{*}{5.41$\uparrow$} & 25.66 & \multirow{2}{*}{8.46$\uparrow$} \\ 
DRPC~\cite{yue2019domain} & \textbf{37.42} & & 32.14 &  & 34.12 &     \\ 
\midrule
Baseline & 28.95 & \multirow{2}{*}{\textbf{7.63}$\uparrow$} & 25.14 & \multirow{2}{*}{\textbf{10.06}$\uparrow$} & 28.18  & \multirow{2}{*}{\textbf{12.15}$\uparrow$}  \\ 
Ours (ISW)   & 36.58 & & \textbf{35.20} &  & \textbf{40.33} & \\ 
% \toprule
% Baseline            & 21.7 & 22.3 & \multirow{2}{*}{N/A} & \multirow{2}{*}{15.2} \\ 
% CyCADA~\cite{hoffman2018cycada} & 38.7 & 35.7 &            & \\ 
\bottomrule
\end{tabular}
\end{center}
\vspace*{-0.15cm}
\caption{mIoU(\%) comparison with IBN-Net and DRPC trained on GTAV train set. The backbone is ResNet-50. Note that IBN-Net does not report the performance on BDD-100K and Mapillary.}
\label{tab_best_model_8}
\vspace*{-0.2cm}
\end{table}
% \vspace{-0.2cm}

\vspace{-0.35cm}
\subsubsection{Comparison with other DG and DA methods}
\vspace{-0.2cm}
This subsection compares our method with two existing DG methods on semantic segmentation task, based on the results reported in the papers~\cite{pan2018two,yue2019domain}. DRPC~\cite{yue2019domain} proposes a domain randomization method, which maps the synthetic images to multiple auxiliary real domains using image-to-image translation with the style of real images (\textit{e.g.,} ImageNet). As shown in Table~\ref{tab_best_model_8}, our model gains the largest performance increase on average, compared to other methods such as IBN-Net~\cite{pan2018two} and DRPC~\cite{yue2019domain}. Our method shows a large amount of performance improvement on BDD-100K and Mapillary datasets that involve significantly more diverse driving scenes than Cityscapes.

In addition, we compare the result of our method with those reported from several domain adaptation methods. See the supplementary Section \todow{A.1}.

\begin{table}[t!]
\vspace*{-0.0cm}
% \hspace*{-0.2cm}
\begin{center}
\setlength\tabcolsep{4.2pt}
\footnotesize
\begin{tabular}{c|c|c|c}
\toprule
Models & \# of Params & GFLOPS & Inference Time (ms) \\
\drule
Baseline  & 45.082M & 554.31 &  10.48\\ 
\midrule
$^\dagger$IBN-Net~\cite{pan2018two}      & 45.083M & 554.31 &  10.51\\
% $^\dagger$IBN-Net~\cite{pan2018two}      & 45.083M & 554.31 &  9.03 \\ 
\midrule
$^\dagger$IterNorm~\cite{huang2019iterative} & 45.081M & 554.31 &  40.31\\ 
% $^\dagger$IterNorm~\cite{huang2019iterative} & 45.081M & 554.31 &  45.44 \\ 
\midrule
Ours    & 45.081M & 554.31 &  10.43 \\
% Ours    & 45.081M & 554.31 &  9.14 \\
\bottomrule
\end{tabular}
\end{center}
\vspace*{-0.2cm}
\caption{Comparison of computational cost. Tested with the image size of 2048$\times$1024 on NVIDIA A100 GPU. The inference time is averaged over 500 trials. $^\dagger$ denotes re-implemented models.}
\label{tab_computational_cost}
\vspace*{-0.5cm}
\end{table}

\vspace{-0.3cm}
\subsubsection{Computational cost analysis}
\vspace{-0.15cm}
To ensure our method requires no additional computational cost, we report the number of parameters, GFLOPS, and inference time. As seen in Fig.~\ref{fig:architectural_comparison}, all the models in Table~\ref{tab_computational_cost} share the same network architecture, but with different normalization methods.
% Our method performs instance standardization at the end of the first three convolution layers. In similar manner, IBN-Net~\cite{pan2018two} has instance normalization layers at the same stages. $^\dagger$IterNorm~\cite{huang2019iterative} has whitening transform matrix approximation steps using Netwon's method. As reported in IterNorm~\cite{huang2019iterative},
%this step has similar computation cost to the 3$\times$3 convolution operation. 
As shown in Table~\ref{tab_computational_cost}, our approach performs a whitening transformation without additional computational cost.

\vspace*{-0.0cm}
\subsection{Qualitative Analysis}
\begin{figure}[b!]
\vspace*{-0.3cm}
\begin{center}
  \includegraphics[width=0.98\linewidth]{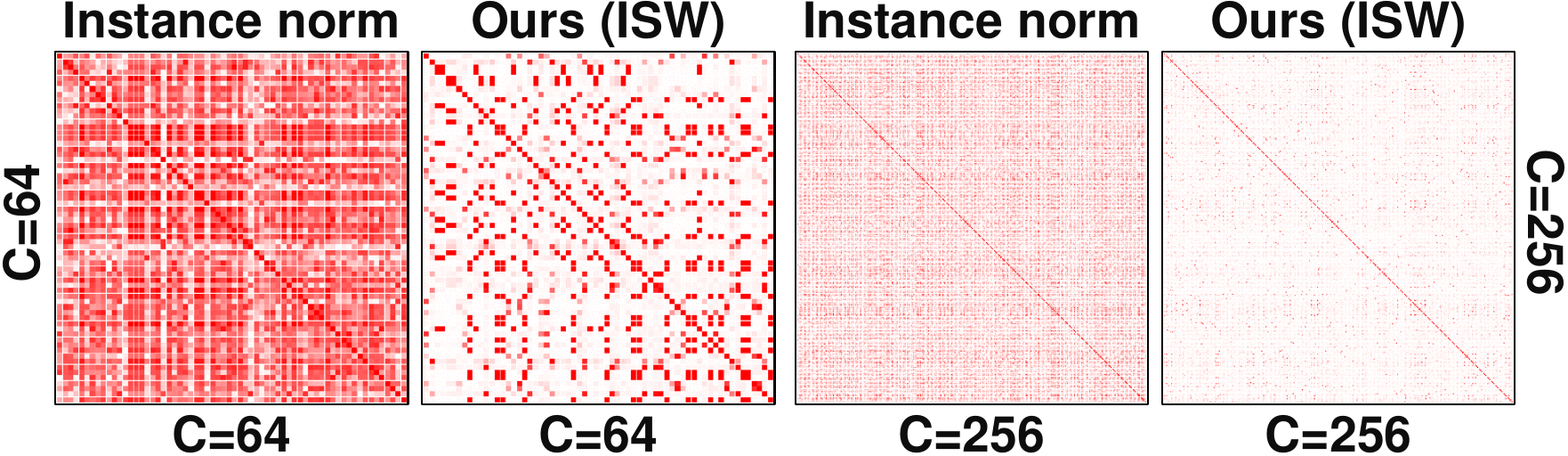}
\end{center}
\vspace*{-0.2cm}
   \caption{Visualization of covariance matrices extracted from IBN-Net and our model. The first and the second pairs are extracted from the first and the second convolution layers, respectively.}
%   the order of the channels are changed by clustering.}
%   Y 축은 height 이고 X축은 채널입니다. 즉 각 vertical position 마다 어떤 channel 을 attention 하고 있는지를 나타냄. Note that 여기서의 채널 순서는 보다 나은 보기를 위해서, 비슷한 채널 들끼리 clustering 하였습니다.}
\label{fig:covariance_matrix}
\vspace*{-0.0cm}
\end{figure}

% This subsection presents the the visualization of the covariance matrix and show
% qualitative analysis of our method. 
\vspace*{-0.05cm}
\paragraph{Comparison of covariance matrices} To show how the covariance matrix is selectively whitened, we visualize the covariance matrix of intermediate feature maps from IBN-Net~\cite{pan2018two} and our model with ISW. As shown in Fig.~\ref{fig:covariance_matrix}, the first pair of covariance matrices are from the first convolution layer and the others are from the second convolution layer. Note that the style information mainly exists in the early layers of the network as pointed out in IBN-Net. Moreover, the style information is encoded as a form of the features covariance as revealed in previous studies~\cite{gatys2015texture, gatys2016image}. Hence, the covariance matrices are sparser at the second pair, compared to the first ones. By comparing the covariance maps from IBN-Net and ISW, we can find the ones from ours are whitened but a small number of covariance elements remain large, showing our ISW selectively eliminates the covariance.

\begin{figure}[t!]
\begin{center}
  \includegraphics[width=1.0\linewidth]{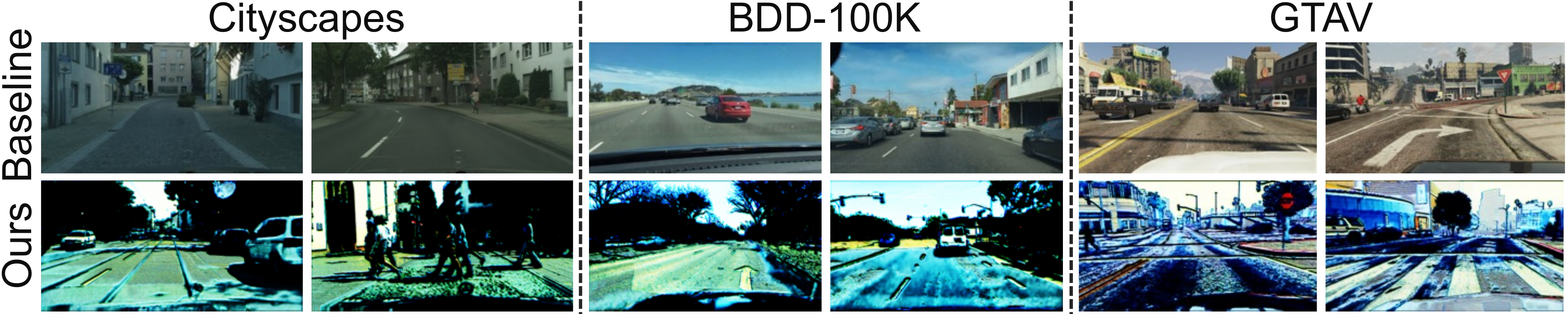}
\end{center}
\vspace*{-0.3cm}
   \caption{Reconstructed images from ISW-whitened feature maps using U-Net; the first row: a baseline backbone, the second row: an ISW model backbone. The image contents are properly maintained while the style such as illumination and colors vanish.}
%   the order of the channels are changed by clustering.}
%   Y 축은 height 이고 X축은 채널입니다. 즉 각 vertical position 마다 어떤 channel 을 attention 하고 있는지를 나타냄. Note that 여기서의 채널 순서는 보다 나은 보기를 위해서, 비슷한 채널 들끼리 clustering 하였습니다.}
\label{fig:reconstruction}
\vspace*{-0.45cm}
\end{figure}

\vspace*{-0.45cm}
\paragraph{Reconstructing images with whitened features} For in-depth analysis, we reconstruct input images from the whitened feature maps of our ISW model. For the experiment, we adopt U-Net~\cite{ronneberger2015u} as reconstruction networks. To newly train a decoder, we append the decoder to the backbone of a pre-trained baseline and train the decoder. We then replace the backbone network with the pre-trained ISW model. As seen in Fig.~\ref{fig:reconstruction}, generated images preserve the relevant content information for segmentation while the style information such as illumination and colors is suppressed. These examples support the validity of our approach that selectively suppresses the style information.

%Additional qualitative examples are presented in the supplementary section.

\vspace{-0.1cm}
%------------------------------------------------------------------------
\section{Discussions}
\vspace{-0.1cm}
% \choi{Affine transform 추가하지 않았음. 실험적으로 좋은 결과 나오지 않음. Photometic augmentation 에 대해서 더 exploration 할 필요 있음 그러나 content 를 해치지 않는 최소한의 augmentation 만 적용했음. Batch whitening 도 적용해보았음. 그러나 실험적으로 좋지 않았음}
% \tr{이 내용 discussion으로 따로 섹션 안만들고 conclusion에 포함시키면 좋을 것 같아요! experimental result에도 각 결과별 discussion이 있으니까 얼핏 보면 디스커션이 내용이 없어보일 듯 해서요}
% \sh{Draft입니다. Overclaim 및 지나치게 치부를 드러냈을 가능성이 있네요}
In this section, we discuss potential issues and improvements of our approach for further research. \vspace{-10pt}
\vspace{-0.1cm}
\paragraph{Affine parameters.} Most of the normalization layers contain affine parameters to recover the original distribution and enhance the representation of a network. We attempted to deploy this by adding affine parameters or a 1$\times$1 convolution layer after the normalization layer incorporating our proposed whitening loss. Despite our effort, this approach did not improve our method. We conjecture it is because affine parameters or a 1$\times$1 convolution layer do not have sufficient complexity in recovering the original distribution.\vspace{-20pt}
\vspace{-0.1cm}
\paragraph{Photometric transformation.} Our method adopted photometric transformation to separate the style and content information, where we found that applying color transform and Gaussian blur does not harm the content information. We expect our approach can be further improved by exploring various photometric augmentation techniques.

\vspace{-0.1cm}
%------------------------------------------------------------------------
\section{Conclusions}
\vspace{-0.1cm}
This paper proposed a novel instance selective whitening (ISW) loss, which facilitates disentangling the covariances of the intermediate features into the style- and content-related ones and suppressing only the former to learn the domain-invariant feature representation.
We focused on solving the domain generalization problem in urban-scene segmentation, which has practical impact when deployed in the wild but has not been studied much. 
In this regard, we strive to promote the importance of the domain generalization and inspire new research paths in this area.
%This paper proposed a novel instance selective whitening (ISW) loss for improving the domain generalization ability of deep neural networks for semantic segmentation, which facilitates disentangling the covariances of the intermediate features into the style- and content-related ones and suppressing only the former to learn the domain-invariant feature representation.
%Through the visualization and analysis, we have discovered meaningful interpretations of the covariance matrix and effects of the instance whitening procedure and showed how we can leverage the feature covariance. 
%Extensive experiments using a variety of benchmark datasets demonstrated the superiority of our method for domain generalization in semantic segmentation tasks. %As future work, we plan to extend our proposed approach to other domain generalization tasks.

\vspace*{0.15cm}
{
\small
\noindent\textbf{Acknowledgments}
This work was partially supported by Institute of Information \& communications Technology Planning \& Evaluation (IITP) grant funded by the Korea government(MSIT) (No. 2019-0-00075, Artificial Intelligence Graduate School Program(KAIST) and No. 2020-0-00368, A Neural-Symbolic Model for Knowledge Acquisition and Inference Techniques), the National Research Foundation of Korea (NRF) grant funded by the Korean government (MSIT) (No. NRF-2019R1A2C4070420).
}
% This work was partially supported by Institute of Information \& communications Technology Planning \& Evaluation (IITP) grant funded by the Korea government (MSIT) (No.2019-0-00075, Artificial Intelligence Graduate School Program (KAIST)), the National Research Foundation of Korea (NRF) grant funded by the Korean government (MSIP) (No. NRF-2018M3E3A1057305).

\clearpage
{\small
\bibliographystyle{latex/ieee_fullname}
\bibliography{egbib}
}
\clearpage
\def\thesection{\Alph{section}}
\setcounter{section}{0}
%%%%%%%%% BODY TEXT
% \setcounter{section}{5}
\section{Supplementary Material} \label{supple}
\vspace{-0.1cm}
This supplementary section provides additional quantitative results to examine hyper-parameter impacts, further implementation details, and qualitative results. %Our code and pretrained models are available at \url{https://github.com/shachoi/RobustNet}. These resources will be publicly available at GitHub in the near future.

\begin{figure}[h!]
\vspace*{-0.15cm}
  \centering\includegraphics[width=1.0\linewidth]{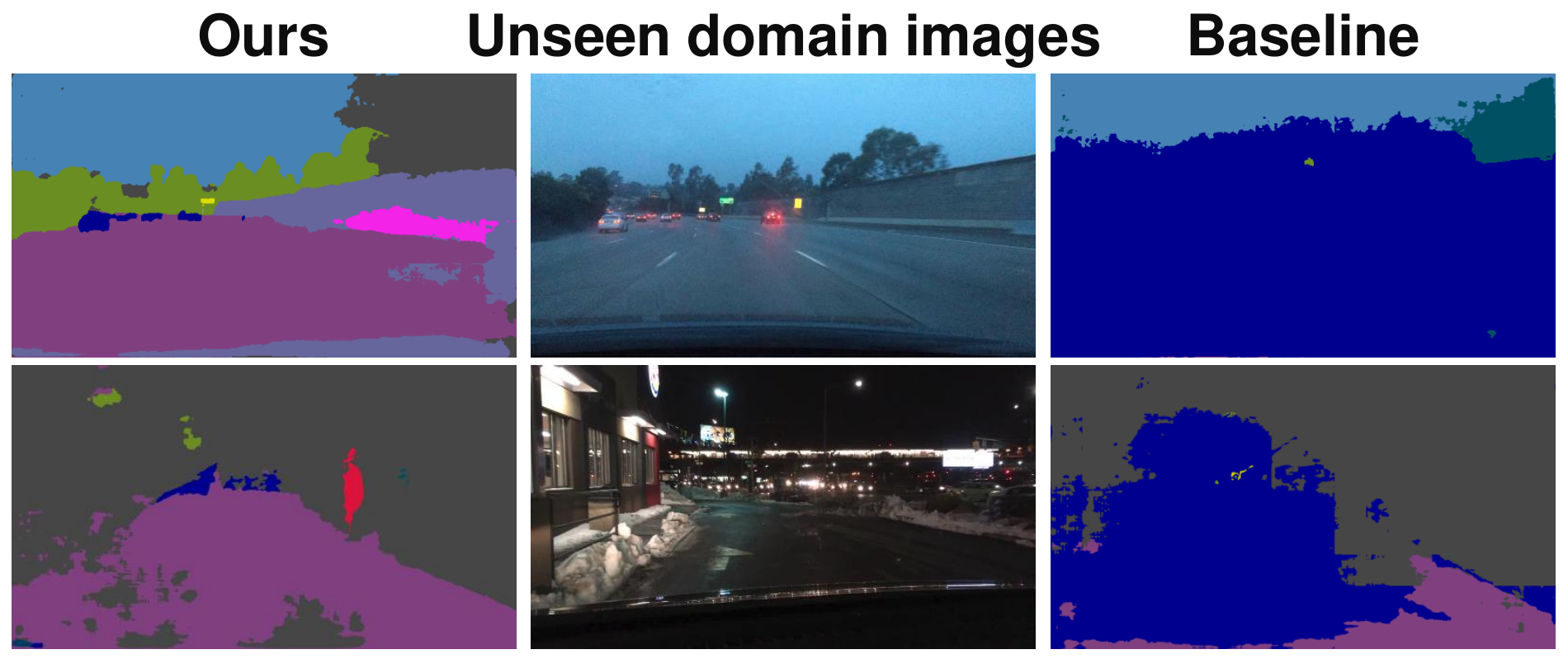}
  \vspace*{-0.2cm}
  \caption{Segmentation results on BDD-100K with the models trained on Cityscapes. The upper image contains dust and water drops on the windshield, and the lower one has an extreme domain shift (\textit{i.e.}, night and snow). Note that Cityscapes does not contain any images taken at night or under a snow condition.
  }
\label{fig:seg_map}
\vspace*{-0.3cm}
\end{figure}

Comparison of segmentation results is shown in Fig.~\ref{fig:seg_map}. Our method makes reasonable predictions, while the baseline completely fails on them.

\vspace{-0.0cm}
\subsection{Comparison with DA methods}
\begin{figure}[h]
\centering
\begin{center}
  \includegraphics[width=1.0\linewidth]{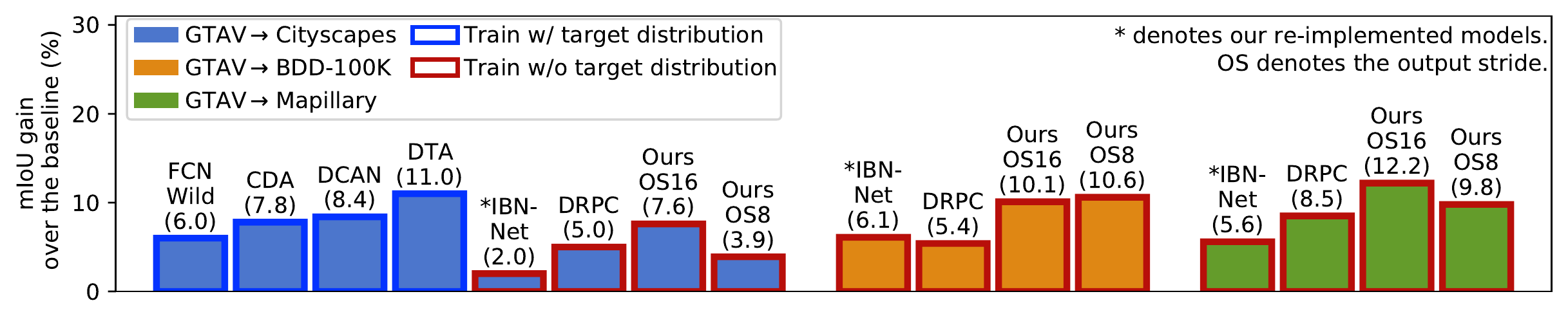}
\end{center}
\vspace*{-0.3cm}
   \caption{Comparison of mIoU gain($\%$) from the baseline for each method. Other methods compared to ours are FCN Wild~\cite{hoffman2016fcns}, CDA~\cite{zhang2017curriculum}, DCAN~\cite{wu2018dcan}, DTA~\cite{lee2019drop}, IBN-Net~\cite{pan2018two}, and DRPC~\cite{zhao2017pyramid}.}
%   the order of the channels are changed by clustering.}
%   Y 축은 height 이고 X축은 채널입니다. 즉 각 vertical position 마다 어떤 channel 을 attention 하고 있는지를 나타냄. Note that 여기서의 채널 순서는 보다 나은 보기를 위해서, 비슷한 채널 들끼리 clustering 하였습니다.}
\label{fig:da_comparison}
\vspace*{-0.2cm}
\end{figure}

\vspace{-0.05cm}
We compare the result of our method with those reported from several domain adaptation (DA) methods under various settings. Fig.~\ref{fig:da_comparison} shows the increase in mIoU from the baseline for each method. Although our method may not be the top performer, it shows comparable results to other DA methods. Note that DA methods require access to the target domain to solve DA problems. In contrast, our method is designed to improve generalization performance on an arbitrary \emph{unseen} domain under the assumption of no access to the target domain, so we believe a comparison with DA methods under the same setting is impossible. However, we expect to solve DA by extending our key idea of \emph{selectively} removing style-sensitive covariances to \emph{selectively} matching such covariances between source and target domain.
%Additionally, our method can work as a complement to various DA methods.
% Additionally, our method can be employed with other DA methods at the same time and would be complementary with them.
% As future work, we plan to extend our proposed methods to DA tasks.

\begin{table}[b!]
\vspace{-0.35cm}
\begin{center}
\footnotesize
\begin{tabular}{c|c|c|c|c||c}
\toprule
Models (GTAV) & C & B & M & S & G\\
\drule
Baseline & 28.95 & 25.14 & 28.18 & 26.23 & \textbf{73.45}  \\ 
\midrule
Ours (ISW), $k$=2 & 35.46 & 35.00 & 39.38 & 27.70 & 72.08 \\ 
\midrule
% Ours (ISW) $k$=5 &  50.52   &   57.70     &    45.28   &    25.37   &     76.49  \\ 
% \midrule
Ours (ISW), $k$=3 & \textbf{36.58} & \textbf{35.20} & \textbf{40.33} & \textbf{28.30} & 72.10 \\ 
\midrule
Ours (ISW), $k$=5 & 34.84 & 33.58 & 39.25 & 27.52 & 72.31 \\ 
\midrule
Ours (ISW), $k$=10 & 33.58 & 33.76 & 38.96 & 27.68 & 72.24 \\ 
\midrule
% $k$=75, $m$=74 & 50.25 &  57.65   &  45.08  & 25.36  & 76.48       \\ 
% \midrule
Ours (ISW), $k$=20 & 33.66 &  33.29 & 38.70  &  27.47  &  72.10     \\ 
\midrule
Ours (IW) & 33.21 & 32.67 & 37.35 & 27.57 & 72.06 \\ 
\bottomrule
\end{tabular}
\end{center}
\vspace*{-0.0cm}
\caption{Comparison of mIoU(\%) on five different validation sets according to $k$ value. Cityscapes (C), BDD-100K (B), Mapillary (M), SYNTHIA (S), and GTAV (G). The models are trained on GTAV. ResNet-50 is adopted, and an output stride of 16 is used. $^\dagger$ denotes re-implemented models. These experiments are conducted three times, and the average results are reported.}
\label{tab_hyper_parameters1}
\vspace{-0.0cm}
\end{table}

% \vspace{-0.2cm}
\begin{figure}[t!]
\vspace{-0.1cm}
\centering
  \includegraphics[width=0.98\linewidth]{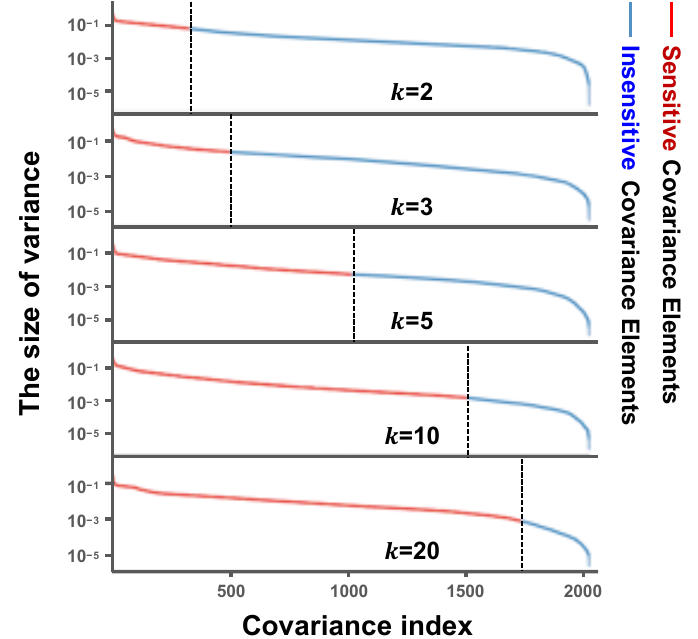}
  \vspace*{-0.0cm}
  \caption{The curves denote the magnitude of the variance of each covariance element across the photometric transformations. The vertical dashed lines represent the threshold to separate the covariance elements. The magnitudes of the variance are extracted from the covariance matrix calculated in the input convolutional layer. The y-axis is in log-scale.}
  %\tr{그림에도 "Magnitude of variance"로 수정이 가능할까요? 그리고 가능하다면 이미지에 있는 색으로 sensitive, insensitive 글씨 색바꾸면 더 좋을 것 같아요(혹은 그 반대!) }}
\label{fig:supp_cov_clustering}
\vspace*{-0.3cm}
\end{figure}

\vspace{-0.0cm}
\subsection{Hyper-parameter Impacts}
\vspace{-0.05cm}
\paragraph{Criteria for separating covariance elements}
We adopt $k$-means clustering to separate covariance elements into two groups, domain-specific style and domain-invariant content, according to the variance of each covariance element across various photometric transformations such as color jittering and Gaussian blur. As specified in Section \todow{4.3}, after dividing the covariance elements into $k$ clusters by the magnitude of the variance, the clusters from the first to the $m$-th are considered to be insensitive, and the remaining clusters are considered sensitive to photometric transformation. We set $m$ to one and search the optimal $k$ through the hyper-parameter search. Fig.~\ref{fig:supp_cov_clustering} shows the threshold where the covariances are divided into two groups depending on the $k$ value.
Table~\ref{tab_hyper_parameters1} shows the changes in mIoU performance according to the $k$ values, suggesting the optimal $k$ as 3. Also, we can see that ours (ISW) performs better than IBN-Net or ours (IW) for all $k$ values. Note that ours (IW) applies instance whitening loss to all covariance elements, while ours (ISW) applies it to a part of the covariance elements according to the $k$ value.

\paragraph{Margin $\delta$ in instance-relaxed whitening (IRW) loss}
\vspace*{-0.3cm}
As described in Section \todow{4.2}, we propose margin-based relaxation of whitening loss. Table~\ref{tab_hyper_parameters2} shows the performance of ours (IRW) according to the margin $\delta$.

\begin{table}[t!]
\vspace{-0.0cm}
\begin{center}
\footnotesize
\begin{tabular}{c|c|c|c|c||c}
\toprule
Models (GTAV) & C & B & M & S & G \\
\drule
Baseline & 28.95 & 25.14 & 28.18 & 26.23 & \textbf{73.45}  \\ 
\midrule
Ours (IRW), $\delta$=1$/$16 &  32.49 & 32.53 & 37.51 &  \textbf{27.77} & 72.18      \\ 
\midrule
Ours (IRW), $\delta$=1$/$32 & 33.30 & 33.17  &  38.03   &   27.43   &   71.96   \\ 
\midrule
Ours (IRW), $\delta$=1$/$64 & \textbf{33.57} & \textbf{33.18}   & \textbf{38.42}  &  27.29   &  71.96 \\ 
\midrule
Ours (IRW), $\delta$=1$/$128 & 32.85 & 32.40 & 37.36 & 27.43 & 72.21      \\ 
\midrule
Ours (IRW), $\delta$=1$/$256 & 32.45 & 32.32 & 37.93  & 27.48 & 72.12   \\ 
\midrule
Ours (IW) & 33.21 & 32.67 & 37.35 & 27.57 & 72.06 \\ 
\bottomrule
\end{tabular}
\end{center}
\vspace*{-0.2cm}
\caption{Comparison of mIoU(\%) on five different validation sets according to $\delta$ value. The models are trained on GTAV train set. ResNet-50 is adopted and an output stride of 16 is used. These experiments are conducted three times, and the average results are reported.}
\label{tab_hyper_parameters2}
\vspace{-0.2cm}
\end{table}

\paragraph{Weight $\gamma$ of instance-selective whitening (ISW) loss}
\vspace*{-0.3cm}
As described in Section \todow{4.4}, we empirically set the weight $\gamma$ of the proposed ISW loss as 0.6. Table~\ref{tab_hyper_parameters3} shows the impact of changing $\gamma$.

% \todow{ The balance between losses is crucial for optimization and should be determined by experiments.}
\begin{table}[h!]
\vspace{-0.0cm}
\begin{center}
\footnotesize
\begin{tabular}{c|c|c|c|c||c}
\toprule
Models (GTAV) & C & B & M & S & G \\
\drule
Ours (ISW), $\gamma$=0.4 & 35.60 & 34.07 & 38.98 &  28.10 & 71.96      \\ 
\midrule
Ours (ISW), $\gamma$=0.6 & \textbf{36.58} & \textbf{35.20}  & \textbf{40.33}   & \textbf{28.30}   &   \textbf{72.10}   \\ 
\midrule
Ours (ISW), $\gamma$=0.8 & 35.73 & 34.01 & 39.69  &  27.44   &  71.96 \\ 
\bottomrule
\end{tabular}
\end{center}
\vspace*{-0.2cm}
\caption{Comparison of mIoU(\%) on five different validation sets according to $\gamma$ value. The models are trained on GTAV train set. ResNet-50 is adopted and an output stride of 16 is used. These experiments are conducted three times, and the average results are reported.}
\label{tab_hyper_parameters3}
\vspace{-0.1cm}
\end{table}

\vspace*{-0.2cm}
\subsection{Further Implementation Details}
\begin{figure}[h!]
\vspace*{-0.1cm}
\centering
  \includegraphics[width=1.00\linewidth]{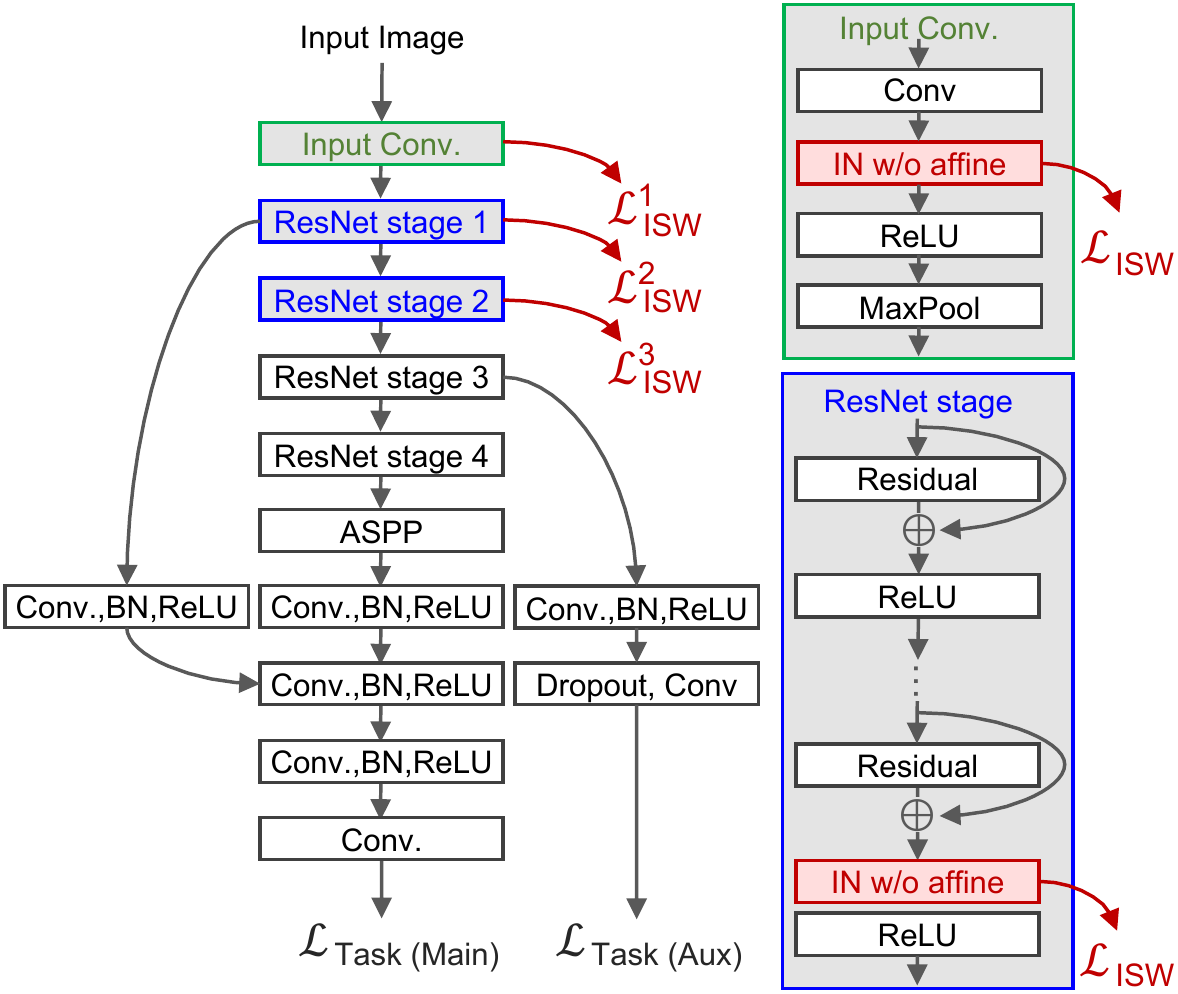}
  \vspace*{-0.2cm}
  \caption{Detailed architecture of the segmentation model.}
  %\tr{정말 매우 사소한건데 Conv뒤에 온점이 빠진 네모칸이 두군데 보여요}}
\label{fig:supp_detailed}
\vspace*{-0.3cm}
\end{figure}

\vspace*{-0.0cm}
Fig.~\ref{fig:supp_detailed} shows the detailed architecture of the semantic segmentation networks based on ResNet and DeepLabV3+. We adopt the auxiliary per-pixel cross-entropy loss proposed in PSPNet~\cite{zhao2017pyramid} and concatenate the low-level features from the ResNet stage 1 to the high-level features according to the encoder-decoder architecture proposed in DeepLabV3+. Instance normalization (IN) with ISW loss replaces batch normalization (BN) 
%\tr{figure 3에는 IN BN이라서 여기에 혹시 싶어서 괄호 추가했어요}
in the input convolutional layer, and these ones are added after the skip-connection of the last residual block for each ResNet stage.
% For ResNet stages, ISW is applied after the last residual connection and for the input convolution layer, which doesn't have residual connection, we add ISW between the convolution layer and ReLU unit.
As IBN-Net~\cite{pan2018two} pointed out, earlier layers tend to encode the style information, hence we only adopt the ISW loss to the input convolutional layer and ResNet stage 1 and 2. In the end, the final loss $\mathcal{L}_{\text{Total}}$ is formulated as,
$$\mathcal{L}_{\text{Total}} = \mathcal{L}_{\text{Task (main)}} + \gamma_1\mathcal{L}_{\text{Task (aux.)}} + \gamma_2(\frac{1}{3}\sum_{i=1}^3\mathcal{L}_{\text{ISW}}^i),$$
where the $\gamma_1$ is 0.4 and the $\gamma_2$ is 0.6. We set the batch size to 8 for Cityscapes and 16 for GTA. For the photometric transformation, we apply Gaussian blur and color jittering implemented in Pytorch with a brightness of 0.8, contrast of 0.8, saturation of 0.8, and hue of 0.3.
% As described in Section 4.3, we apply \textit{k}-means clustering on the variance matrix $\mathbf{V}$ to calculate . However

\begin{figure*}[!b]
\vspace*{-0.2cm}
\centering
  \includegraphics[width=1.0\linewidth]{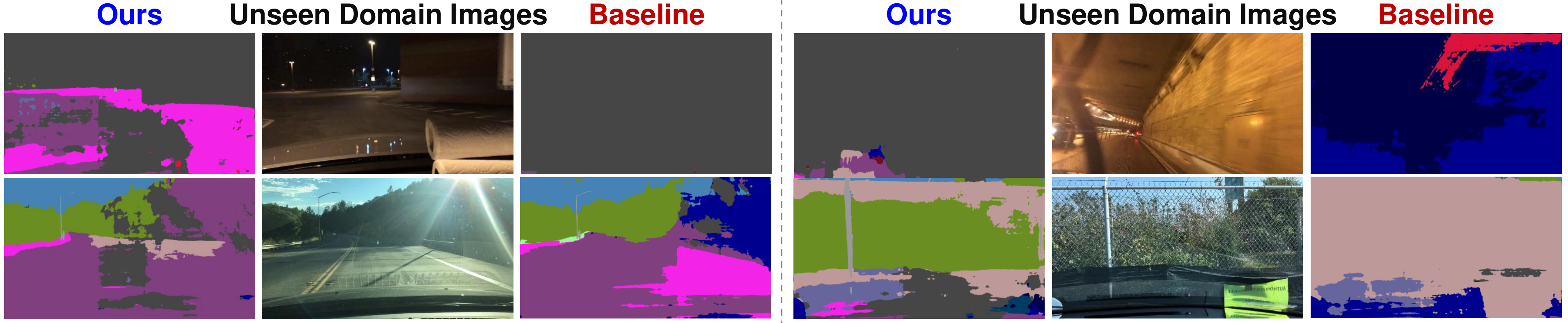}
  %\vspace*{0.05cm}
  \caption{Comparison of failure cases of our method and the baseline.}
\label{fig:supp_failure_results}
% \vspace*{-0.2cm}
\end{figure*}

\subsection{Additional Qualitative Results}
\vspace*{-0.1cm}
This section demonstrates additional qualitative results. We first present the comparison of the segmentation results on a \emph{seen} domain (\textit{i.e.,} Cityscapes) and diverse driving conditions in BDD-100K, and then show the failure cases of our method. Besides, we show the effects of the whitening by comparing the reconstructed images from our proposed approach and the baseline. Finally, we provide the tendency of images from the most sensitive and insensitive covariance elements to the photometric transformation.

\begin{figure*}[!t]
\centering
  \includegraphics[width=1.0\linewidth]{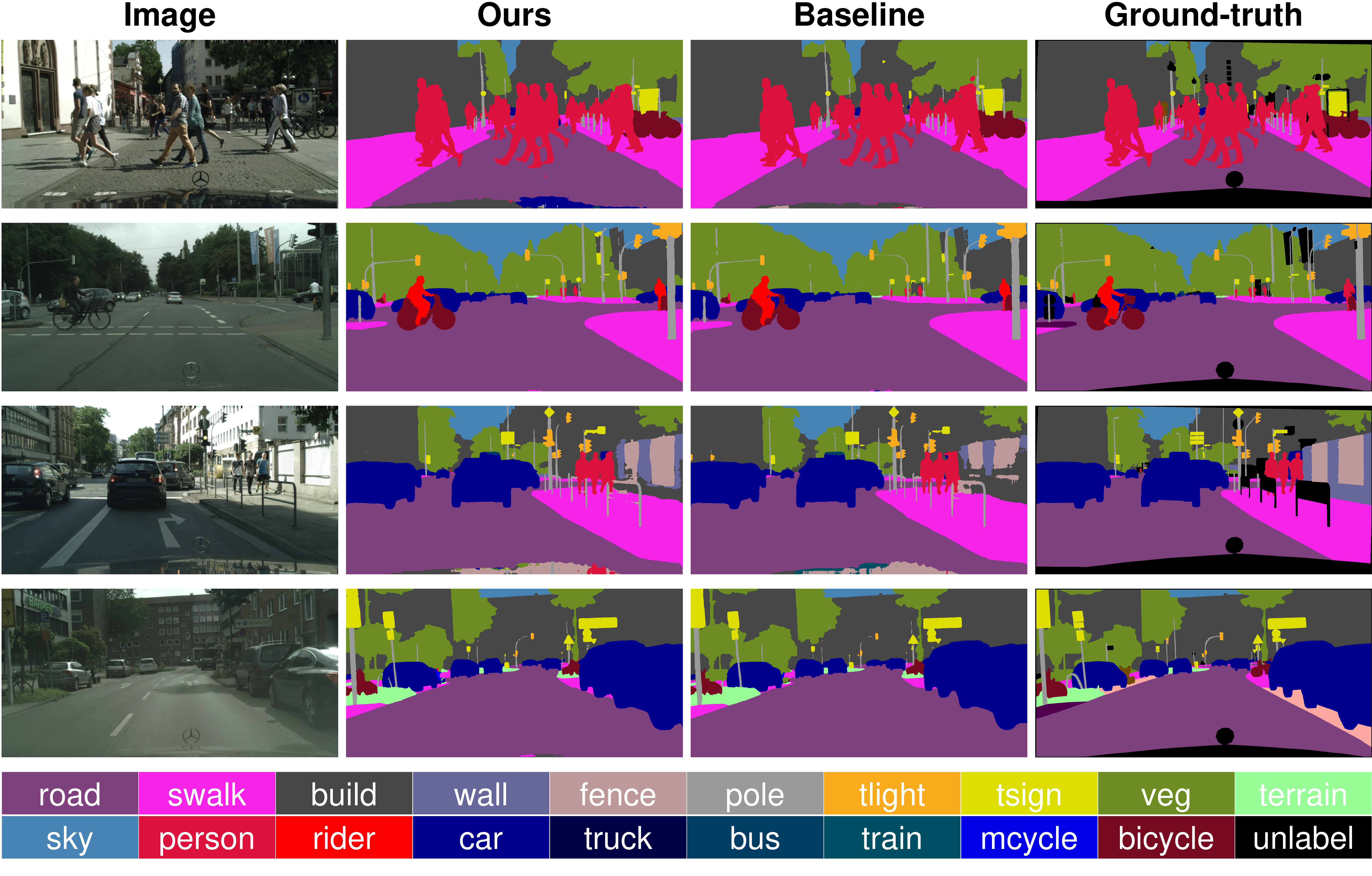}
%   \vspace*{-0.1cm}
  \caption{Segmentation results on \emph{seen} domain images (\textit{i.e.,} Cityscapes).}
\label{fig:supp_seg_result_city}
% \vspace*{-0.3cm}
\end{figure*}

\begin{figure*}[!b]
\vspace*{-0.2cm}
\centering
  \includegraphics[width=1.0\linewidth]{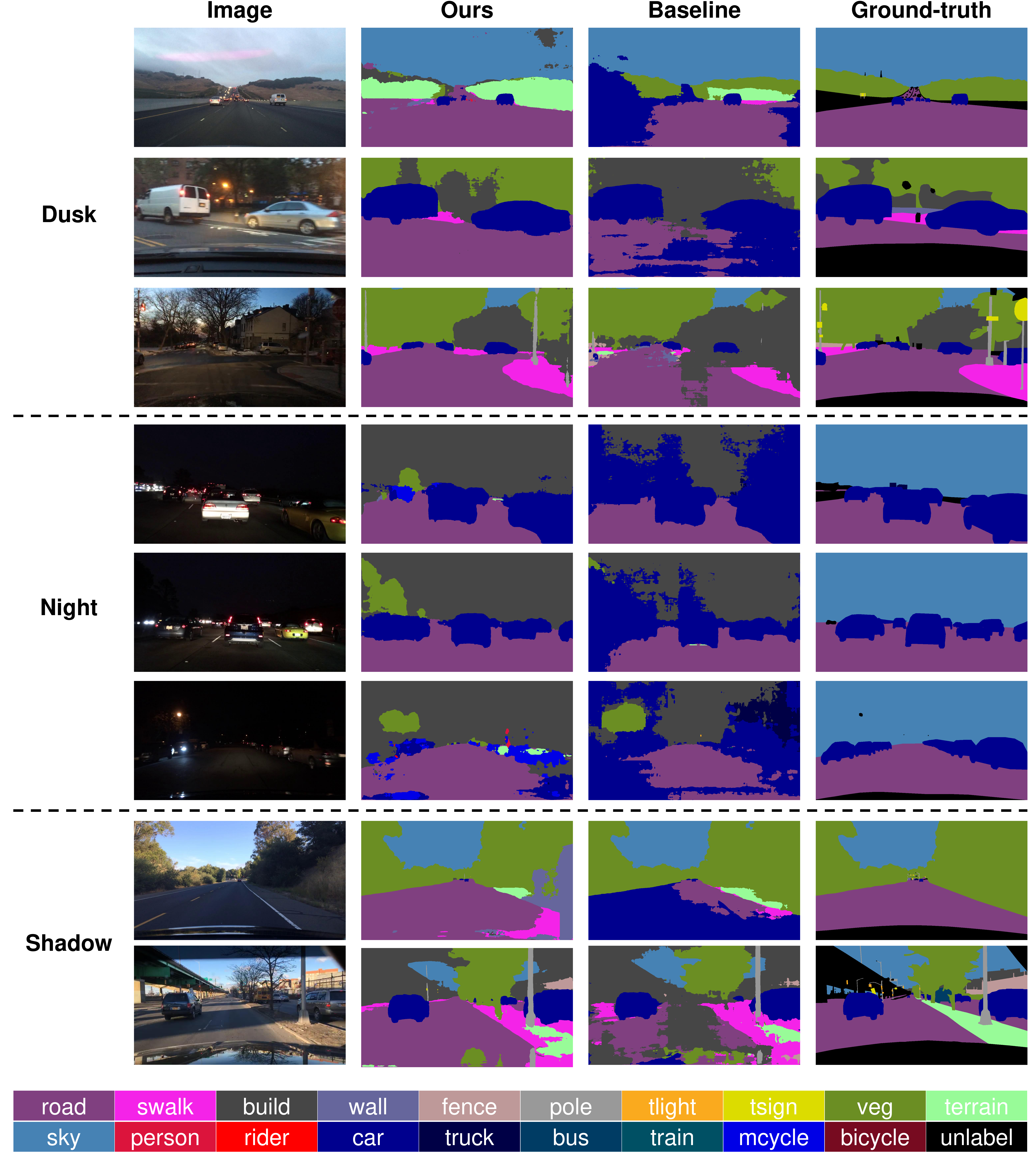}
  \caption{Segmentation results under illumination changes (\textit{i.e.,} dusk, night, and shadow) in BDD-100K with the models trained on Cityscapes.}
\label{fig:supp_seg_result_illumination}
\end{figure*}

\begin{figure*}[!t]
\vspace*{-0.3cm}
\centering
  \includegraphics[width=1.0\linewidth]{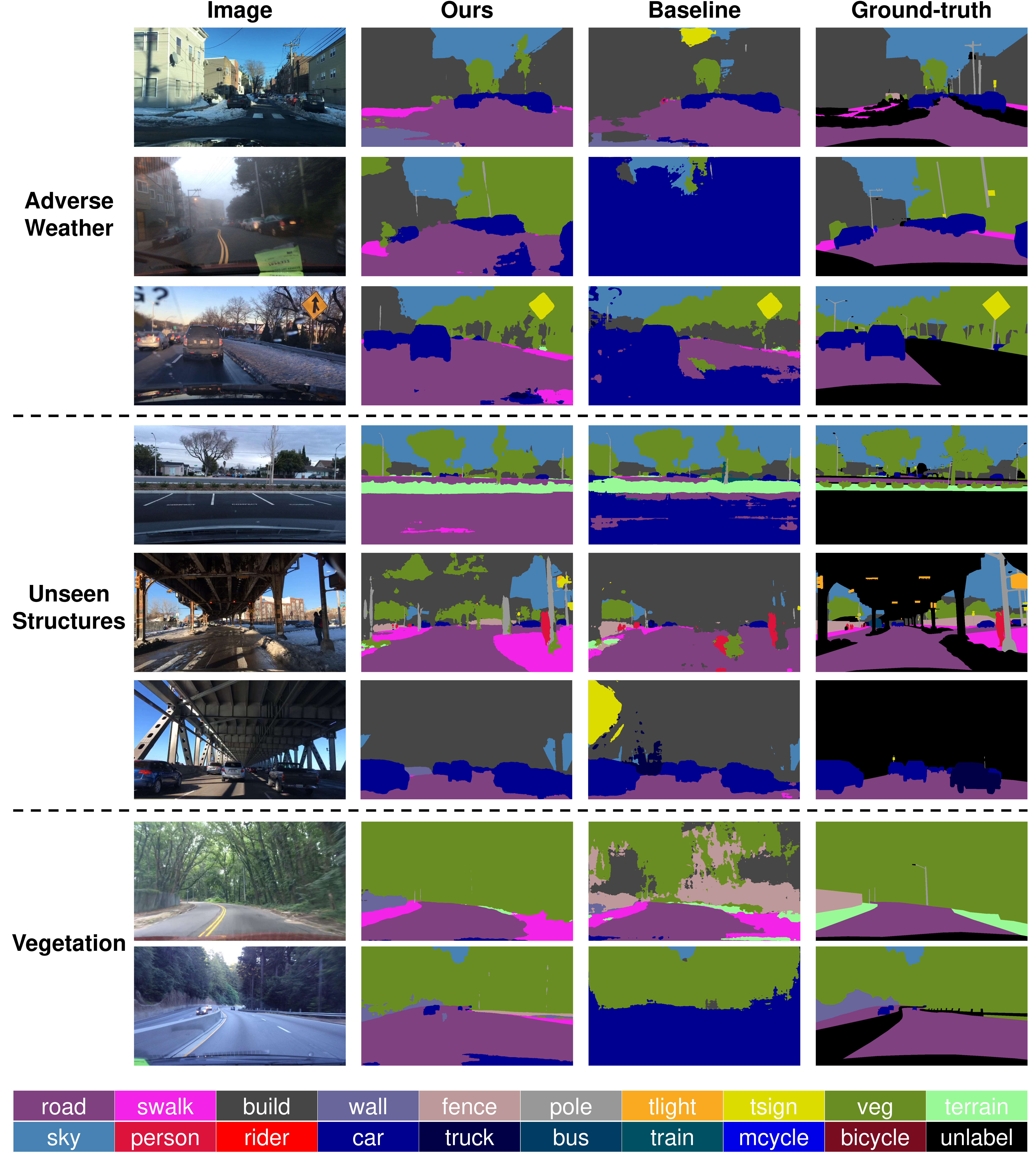}
  \caption{Segmentation results under various circumstances in BDD-100K with the models trained on Cityscapes. Circumstances include adverse weather conditions (\textit{i.e.,} snow and fog), unseen structures (\textit{i.e.,} parking lot and overpass), and vegetation.}
\label{fig:supp_seg_result_various}
\vspace*{-0.3cm}
\end{figure*}

\paragraph{Comparison of segmentation results}
\vspace*{-0.3cm}
To qualitatively describe the effect of our method, we compare the segmentation results from the baseline and ours.
Fig.~\ref{fig:supp_seg_result_city} presents the segmentation results on a \emph{seen} domain (\textit{i.e.,} Cityscapes).
%Both models produce the high quality results.
Similar to the quantitative results reported in Section \todow{5}, even with qualitative results, our model shows comparable performance to the baseline model on the \emph{seen} domain.
%\tr{baseline model에 비해서 comparable하다는 것 맞나요?}
% By this, one can observe that the original performance on a training domain still remains the same with our proposed model.
Fig.~\ref{fig:supp_seg_result_illumination} shows the segmentation results under illumination changes on an \emph{unseen} domain (\textit{i.e.,} BDD-100K). 
Note that Cityscapes dataset only contains images taken at the daytime. 
The first group images are taken at the dusk.
%The first group shows the results on the small but global illumination changes.\choi{small but global?? 이 뭐에요} 
We can see that the baseline model is vulnerable to these changes, but in contrast, our model outputs less damaged maps and reasonably predicts roads and cars.
% One can know the baseline model is susceptible to these changes. In contrast, our model outputs less corrupted maps and predicts the road and cars reasonably.
% In extreme case such as \textit{night}, both models produce the noisy output and fail to predict the sky. However, our method still finds the main components such as the road and cars.
In extreme cases such as at night, both models fail to predict the sky, but our method still finds key components such as roads and cars well. 
In addition, our method produces reasonable segmentation results even for drastic changes in lighting such as shadows, as seen in the third group.
% Moreover, our method produces the reasonable segmentation maps under local illumination changes as described in the third group (\textit{i.e., shadow}). 
Fig.~\ref{fig:supp_seg_result_various} shows the segmentation results under the adverse weather conditions, unseen structures, and lush vegetation.
% To further provide results under diverse driving scenes, we test our method on adverse weather conditions, unseen structures, and vegetation cases, as shown in Fig.~\ref{fig:supp_seg_result_various}. 
% Our model successfully finds the sidewalk partially covered with snow while the baseline fails and marks it as building.
Our model successfully predicts a partially snowy sidewalk, whereas the baseline model incorrectly predicts it as a building.
The second case in the first group shows a foggy urban scene. The baseline fails to cope with these weather changes, while ours still shows fair results. Under the structural changes as shown in the second group, our method finds the road and sidewalk better than the baseline. Moreover, the baseline totally fails to detect the parking lot. In the last case, which is lush vegetation, the baseline produces noisy segmentation results and confused the road as a car. On the other hand, our model shows reasonable performance in both cases. Fig.~\ref{fig:supp_failure_results} shows the failure cases caused by a large domain shift.
% Fig.~\ref{fig:supp_failure_results} shows the failure cases of our method and baseline. If the domain shift goes too huge, both models fail to deal with the shift and generate the totally corrupted segmentation results. 

\begin{figure*}[!t]
\centering
  \includegraphics[width=\linewidth]{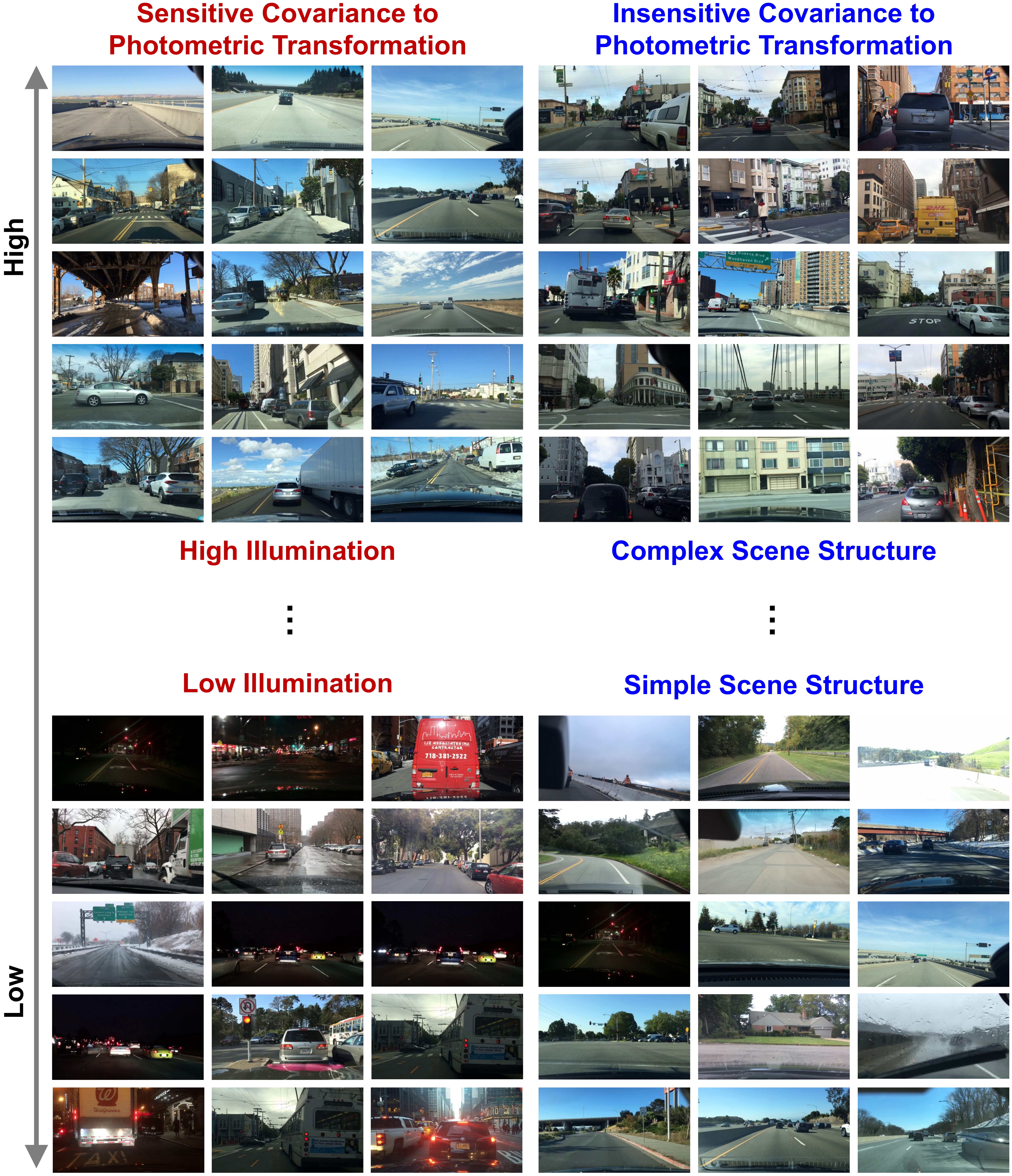}
%   \vspace*{-0.2cm}
  \caption{Tendency of images in BDD-100K dataset along with the covariance changes.}
\label{fig:supp_cov_results}
\vspace*{+1.0cm}
\end{figure*}

\vspace*{-0.4cm}
\paragraph{Covariance effects in images}
%To reveal what covariances are encoding,
To reveal the information that the covariance represents, 
% we first get the indices of the most sensitive and insensitive covariances to the photometric transformation. Then, we sort the BDD-100K images according to the covariances of corresponding indices.
we first identify the most sensitive and insensitive covariances to the photometric transformation. Then, we sort the BDD-100K images according to the magnitude of the identified covariances.
The results are described in Fig.~\ref{fig:supp_cov_results}. In the left group, the images are getting dark as the most sensitive covariance is getting smaller. We conjecture that the corresponding covariance tends to represent the \emph{illumination} information. On the other hand, the right group shows the sorted images along with the most insensitive covariance. The scenes are getting simpler as the covariance gets smaller, which implies that the most insensitive covariance tends to represent the \textit{scene complexity}.

% \vspace*{-0.4cm}
% \paragraph{Interpretation on reconstructed images}
% To further show the effects of our proposed loss, we reconstruct the feature map outputs of our model using U-Net. The results are presented in Fig.~\ref{fig:supp_reconstruction}. The reconstructed images successfully maintain the contents such as objects and edges, while colors and illuminations are washed out.

% \clearpage
% {\small
% \bibliographystyle{latex/ieee_fullname}
% \bibliography{egbib}
% }

% \begin{figure*}[!b]
% \centering
%   \includegraphics[width=\linewidth]{latex/figures/fig_supp_reconstruction.pdf}
%   \caption{Comparison of reconstructed images from the baseline and our method.}
% \label{fig:supp_reconstruction}
% \vspace*{-0.3cm}
% \end{figure*}

% \clearpage
% {\footnotesize
% \bibliographystyle{latex/ieee_fullname}
% \bibliography{egbib}
% }

\end{document}